\documentclass[11pt]{article}
\usepackage{verbatim,amsmath,amssymb}
\usepackage{epsfig,float,color}
\usepackage{epstopdf}
\usepackage{geometry}
\usepackage{setspace}
\usepackage{wrapfig}
\usepackage{hyperref}
\usepackage[utf8]{inputenc}
\usepackage{graphicx}
\usepackage{flushend}
\usepackage[linesnumbered,ruled]{algorithm2e}

\usepackage[font={footnotesize}]{caption}
\usepackage[font={footnotesize}]{subcaption}
\usepackage{footnote}
\usepackage{multicol}
\usepackage{multirow}
\usepackage{enumitem}   
\usepackage{soul}
\usepackage{framed}
\usepackage[titletoc,title]{appendix}
\usepackage{bm}
\usepackage{siunitx}
\usepackage{cite}
\usepackage{booktabs}
\usepackage[T1]{fontenc}
\usepackage{textcomp}
\usepackage[makeroom]{cancel} 
\usepackage{hyperref}
\definecolor{darkblue}{rgb}{0,0,1}
\hypersetup{pdftex=true, colorlinks=true, breaklinks=true, linkcolor=magenta, menucolor=magenta, citecolor=magenta, urlcolor=magenta}

\def\BibTeX{{\rm B\kern-.05em{\sc i\kern-.025em b}\kern-.08em
		T\kern-.1667em\lower.7ex\hbox{E}\kern-.125emX}}
\usepackage{balance}
\usepackage{hyperref}
\definecolor{beaublue}{rgb}{0.94, 0.97, 1.0}
\definecolor{darkblue}{rgb}{0, 0, 1}
\definecolor{LightCyan}{rgb}{0.88,1,1}
\definecolor{mygreen}{RGB}{28,172,0}
\definecolor{mylilas}{RGB}{170,55,241}
\definecolor{grayblue}{RGB}{220,230,240}
\definecolor{topressbg}{rgb}{0.95,0.95,0.92}
\definecolor{topressgreen}{rgb}{0,0.5,0}
\definecolor{topressgray}{rgb}{0.5,0.5,0.5}
\definecolor{topresspurple}{rgb}{0.58,0,0.82}
\definecolor{topressblue}{rgb}{0,0,1}
\hypersetup{
	pdftex=true,
	colorlinks=true,
	breaklinks=true,
	linkcolor=darkblue,
	menucolor=darkblue,
	pagecolor=darkblue,
	citecolor=darkblue,
	urlcolor=darkblue
}
\usepackage{pgfplots}
\usetikzlibrary{positioning, arrows.meta}
\tikzset{%
	myarrow/.style = {-Stealth, shorten >=5pt}
}

\definecolor{darkblue}{rgb}{0,0,1}
\hypersetup{pdftex=true, colorlinks=true, breaklinks=true, linkcolor=darkblue, menucolor=darkblue, pagecolor=darkblue, citecolor=darkblue, urlcolor=darkblue}

\usepackage{array}
\usepackage{mwe}
\newcolumntype{C}[1]{>{\centering\arraybackslash}m{#1}}

\begin{document}
	
	\begin{center}
		\Large{\bf{TiBCLaG: Hybrid design approach for a trigger-induced bistable compliant laparoscopic grasper}}\\
		
	\end{center}
	
	\begin{center}
		Joel J Nellikkunnel$^*$, Prabhat Kumar$^{*,}$$\footnote{pkumar@mae.iith.ac.in}$
		
		\vspace{4mm}
		\small{$*$\textit{Department of Mechanical and Aerospace Engineering, Indian Institute of Technology Hyderabad, Telangana 502285, India}}
		
			Published\footnote{This pdf is the personal version of an article whose final publication is available at \href{https://doi.org/10.1115/1.4072356}{https://asmedigitalcollection.asme.org/medicaldevices}} 
		in \textit {Journal of Medical Devices}, 
		\href{https://doi.org/10.1115/1.4072356}{DOI:10.1115/1.4072356} \\
		Submitted on 06~April 2026, Revised on 12 July 2026, Accepted on 13 July 2026
	\end{center}
	
	\vspace{3mm}
	\rule{\linewidth}{.15mm}
	
	\begin{center}
		\textbf{\large Graphical Abstract}
	\end{center}
	
	\begin{figure}[H]
		\centering
		 \includegraphics[scale =2]{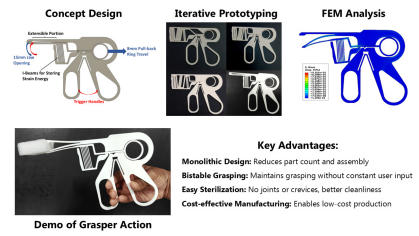}
	\end{figure}
	
	\rule{\linewidth}{0.15mm}
	
	{\bf Abstract:}
Industrial laparoscopic graspers rely on multi-link rigid mechanisms manufactured to tight tolerances, resulting in high manufacturing and assembly costs. This work presents the design and proof-of-concept validation of a monolithic, fully compliant, bistable, laparoscopic grasper that eliminates the need for multiple rigid links, thereby reducing part count. The device integrates a compliant trigger and a compliant gripper end-effector, coupled via a control push-rod, to achieve stable grasping without continuous user input. A hybrid approach combining an analytical model, iterative prototyping, and finite element analyses is employed to design the laparoscopic grasper. The trigger mechanism is synthesized using a Two-Element Beam Constraint Model as a design framework to control the deformation and stiffness of I-beam-like elements. This technique enables elastic energy storage while preventing snap-through instability. The end-effector is designed as a compliant gripper to achieve adaptive grasping through elastic deformation.  The laparoscopic design presented here is fabricated using fused deposition 3D printing. A comparative study of the jaws' opening and closing performance is presented using nonlinear finite element analysis and experimentation.  The fabricated prototype demonstrates reliable bistable actuation, confirming the feasibility of such compliant laparoscopic grasper architectures.\\
	

	\vspace{-4mm}
	\rule{\linewidth}{.15mm}
	
 \section{Introduction} \label{sec:intro}
 Minimally invasive surgeries (MIS), also known as minor access surgeries (MAS), are extremely popular due to their advantages over the traditional open surgical procedures~\cite{Paper18}. They reduce operative trauma and the duration of short-term disability, in addition to facilitating a shorter hospital stay~\cite{Paper24}. Due to the small incision serving as the access point, the surgeon cannot use their hands to manipulate tissues or organs; instead, specialized instruments, such as laparoscopic graspers that can fit through such small openings~\cite{Paper6}, must be used to perform the required operation, while external tissue manipulator instruments~\cite{Paper26} are used to manipulate the workspace.
 
 Note that laparoscopic graspers also have several disadvantages. The operation of the instrument is more technical, less efficient, and causes more fatigue in the surgeon, mainly due to the nature of the surgery~\cite{Paper24, Paper4}. Studies have shown that the current design of laparoscopic graspers can lead to neuropraxia after such procedures~\cite{Paper15}. Since the laparoscopic grasper is the only means of physical interaction between the surgeon and the target organ, the force and haptic feedback it provides to the operator are critical. Sjoerdsma et al.~\cite{Paper11} indicated that such graspers can have mechanical efficiencies as low as 8\%, which implies that only 8\% of the applied force is used in grasping. Hence, the force required to operate the grasper and the force needed to hold the tissue cannot be easily differentiated, which can cause tissue trauma. Westebring-van der Putten et al.~\cite{Paper20} also showed that this poor force feedback can cause the tissue to slip out of the grasp of the instrument. Studies have also been conducted comparing inline and pistol grips in laparoscopic graspers~\cite{Paper1, Paper16}, but have not reached any definitive conclusions. Shakeshaft et al.~\cite{Paper17} demonstrated that curved edges on a compliant grasper can moderate the high pressure exerted on the tissue. At the same time, Marucci et al.~\cite{Paper3} highlighted the ability of compliant materials, such as silicone, to reduce peak tissue pressure.
 
 Traditional laparoscopic mechanisms, and even more complex graspers like the work of Bazman et al.~\cite{Paper33}, rely on linkages, cams, springs, pulleys, belts, and so on to convert the motion of the surgeon's fingers to the grasping motion of the jaws~\cite{Paper21}. The use of multiple parts creates the need for tight tolerancing during manufacturing and increased costs during manufacture and assembly. Sterilization is also tricky due to the sheer number of parts. Although an innovative laparoscopic instrument cleaner was presented by Robertson et al. \cite{robertson2025context}, the authors concluded that the significant operator training challenges involved may hinder its commercial and clinical application. This work presents a trigger-induced bistable compliant laparoscopic grasper mechanism as an alternative to the industry standard. Not only does the proposed solution help alleviate the difficulties of manufacturing, assembly, and sterilization, but the bistable design also applies a constant force to the grasped tissue without requiring user input, successfully preventing slipping. 
 
 Previous research on laparoscopic graspers has focused heavily on the compliant gripper end-effector, initiated by Balazs et al.~\cite{Paper9} in 1998. In particular, Herder and co-workers have spearheaded efforts in the design of statically balanced compliant graspers~\cite{Paper12, Paper13, Paper5} that provide a reliable force feedback to the surgeon by balancing the strain energy due to compliance using a negative-stiffness compensation system. Initially, Herder et al.~\cite{Paper19} developed a statically balanced laparoscopic grasper incorporating a rolling-link mechanism between the handle and the jaws to reduce friction and improve force feedback by maintaining a constant force transmission ratio over the working angle range. Segla et al.~\cite{Paper22} further formulated an optimization problem to obtain a force transmission ratio of one for this grasper, and successfully realized the required dimensions. Tolou and Herder~\cite{Paper12} used a mathematical formulation and finite element method (FEM) simulations to design a statically balanced, fully compliant mechanism (CM). Further, a Pseudo-Rigid Body Model was used by Lamers et al.~\cite{Paper13}. Finally, de Lange et al.~\cite{Paper5} used a topology optimization methodology~\cite{kumar2023honeytop90} to design the compliant grasper. Wang and Lan~\cite{Paper34} developed a statically balanced compliant gripper mechanism that exerts a constant force on the gripper object. However, only the gripper end-effector design was provided; coupling with the instrument handle was not demonstrated. Klok et al.~\cite{Paper2} also pursued a balanced, compliant laparoscopic grasper to improve haptic feedback and mechanical efficiency by using magnets to balance the mechanism. Although the results were promising, the large number of complicated parts involved creates assembly constraints. Robotic laparoscopic graspers were also explored, like the cable-driven continuum mechanism-based flexible robotic laparoscope developed by Wang et al. \cite{Paper30}, and the compact robotic manipulator with a gripper end-effector designed by Quaglia et al.~\cite{Paper32}, but they prove to be dramatically costlier than manual laparoscopic surgeries while offering modest advantages compared to the same for numerous procedures~\cite{Paper31}. Galvin et al.~\cite{Galvin2023} explored the design of a novel singel incision, free motion laparoscopic surgical system using three experiments.
 
 Another prominent research area in the field is multifunctional mechanisms. Incorporating multiple types of end effectors into the same laparoscopic grasper can effectively speed up surgeries that involve frequent tool swaps. End effectors that perform cutting (scissors) and grasping were developed by Frecker et al.~\cite{Paper7}, initially using rigid links and later using a compliant mechanism. In Ref.~\cite{Paper10}, they used topology optimization to design a similar end effector. A three-fingered multifunctional grasper was designed by Rosen et al.~\cite{Paper18} to achieve the combined functions of grasping, retracting, and lifting. However, for all the multifunctional graspers, it was noted that not only were the mechanisms very complex, but also the demonstrated capabilities of the combined end effector were significantly less than those of individual end effectors. Finger-based graspers for this application were also explored by O'Hanley et al.~\cite{Paper6} and Cohn et al.~\cite{Paper8}. While the grasping capabilities were improved, Cohn et al.~\cite{Paper8} specifically demonstrated that, due to the mechanism's multiple parts and low force output, pistons were required to actuate it. A fully multimodal, fully compliant mechanism was also developed by Fernandez-Montoya et al. \cite{Paper27}, which allows the surgeon to tune the force transmission ratio of the grasper by varying the flexure's effective length. However, the multimodal behavior of the grasper was observed only in the case of gripping extremely stiff tissue.
 
 Multifunctional mechanisms developed for laparoscopic graspers and finger-based graspers face the inherent issue of complex assemblies, thereby increasing both manufacturing costs due to the tight tolerances required and assembly costs. Moreover, to the best of the authors' knowledge, multifunctional mechanisms have yet to demonstrate greater effectiveness in performing their individual functions compared to a standard unifunctional grasper. Our work aims to eliminate the main costs involved in manufacturing and assembly by developing a monolithic, compliant laparoscopic grasper that can be easily fabricated via fused deposition modeling. Since the research focus in this area has primarily been on the gripper end-effector, we aim to develop a complete compliant mechanism for the grasper, integrating the end-effector with a bistable trigger handle. We call it the \textbf{t}rigger-\textbf{i}nduced monolithic \textbf{b}istable \textbf{c}ompliant \textbf{la}paroscopic \textbf{g}rasper (TiBCLaG). Here, the design methodology combines an analytical beam-constraint model~\cite{Liu2021ABC} with iterative physical prototyping and FEM analyses. This hybrid approach enables geometry estimation for the proposed I-beam system, while prototyping captures multi-beam interaction effects.  The performance of the proposed TiBCLaG is experimentally and numerically verified. A direct experimental test with different weights is also performed to demonstrate the jaw-opening behavior of the proposed 3D-printed TiBCLaG.
 
 \par The remainder of the paper is organized as follows. Section \ref{sec:method} presents the methodology used for the development of the laparoscopic grasper mechanism. Section \ref{sec:results} presents the results obtained after the fabrication of the grasper, as well as its limitations and scope for future work. Lastly, Sec. \ref{sec:conclusion} presents the conclusion.
 
 \section{Methodology} \label{sec:method}
\begin{figure}[h!]
 	\centering\includegraphics[width=1\linewidth]{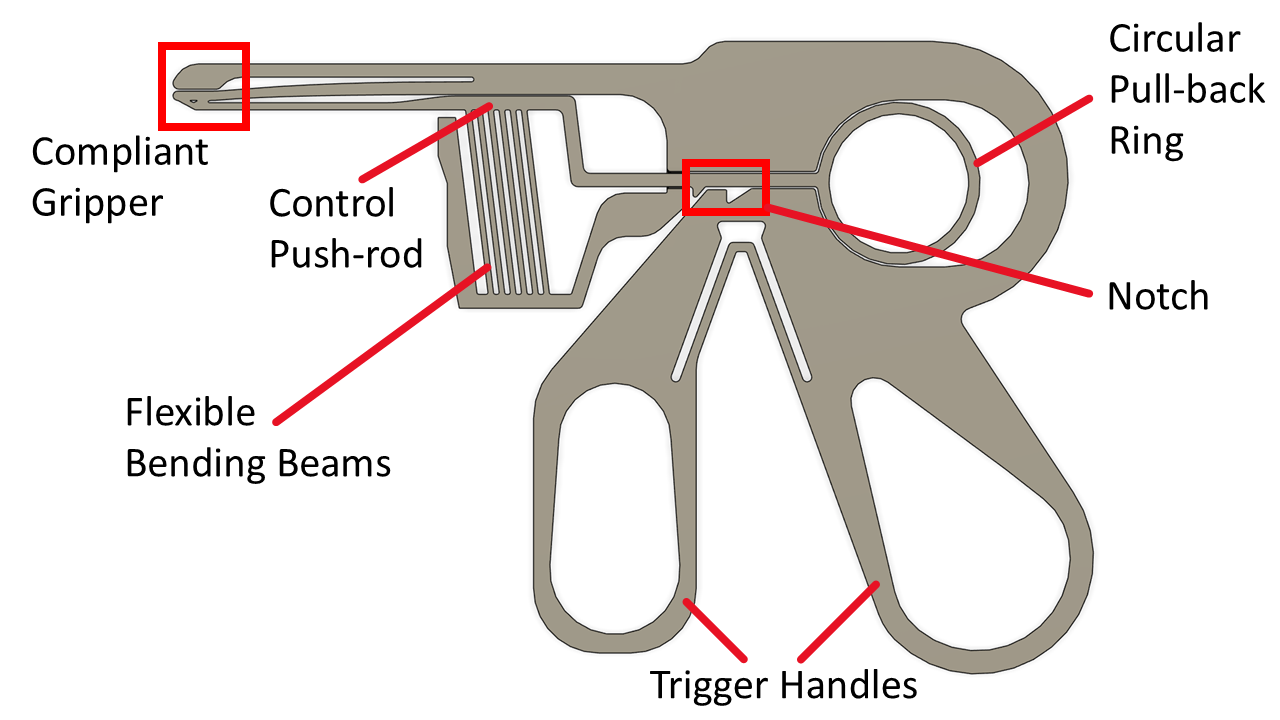}
 	\caption{Plan view of the 3D CAD model of the proposed device. \label{fig:1}}
 \end{figure}

\begin{figure*}
 	\centering
 	\begin{subfigure}{0.3\textwidth}
 		\centering
 		\includegraphics[height=3.7cm,keepaspectratio]{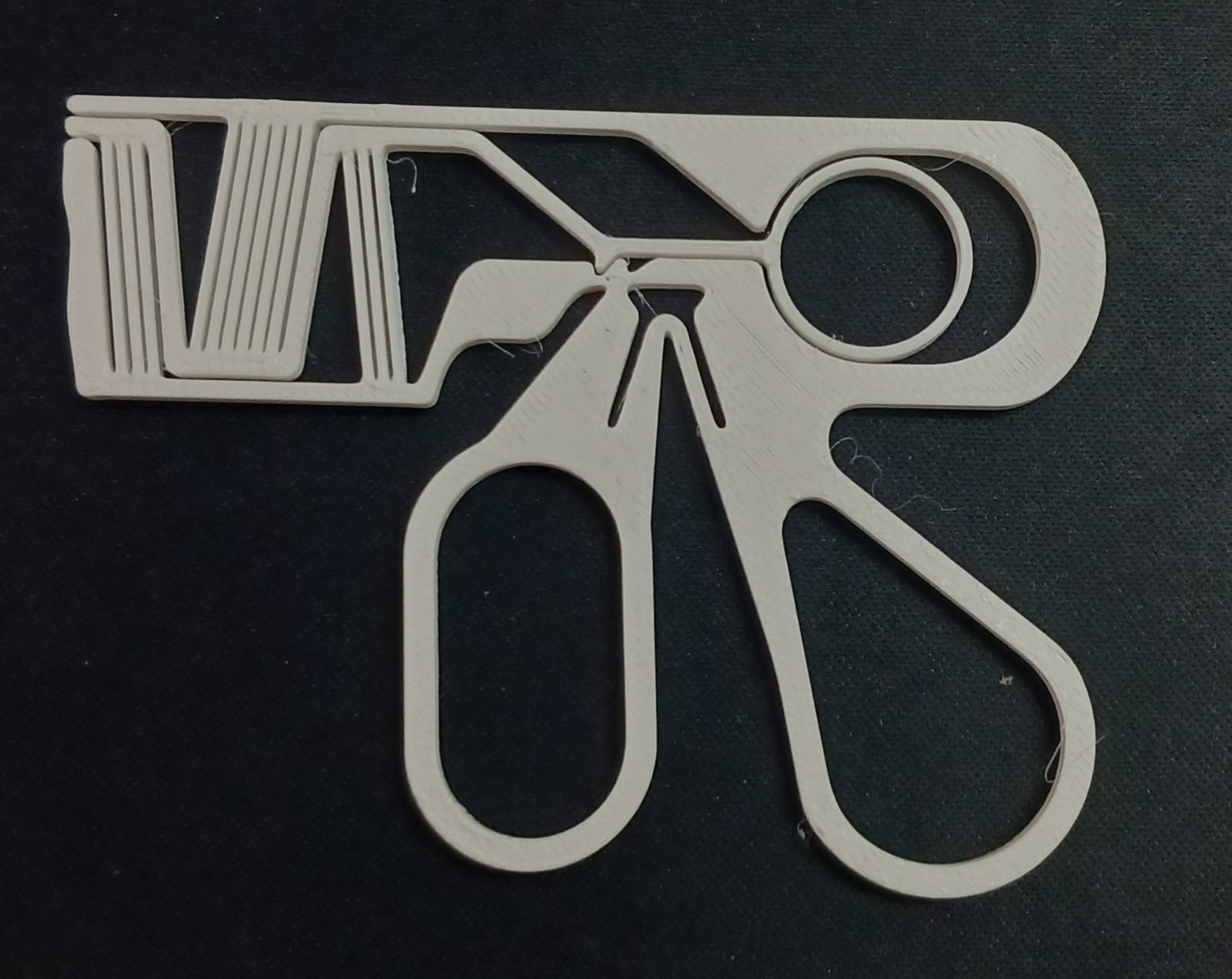}
 		\subcaption{\label{fig:prot-1-rest}}
 	\end{subfigure}
 	\begin{subfigure}{0.3\textwidth}
 		\centering
 		\includegraphics[height=3.7cm,keepaspectratio]{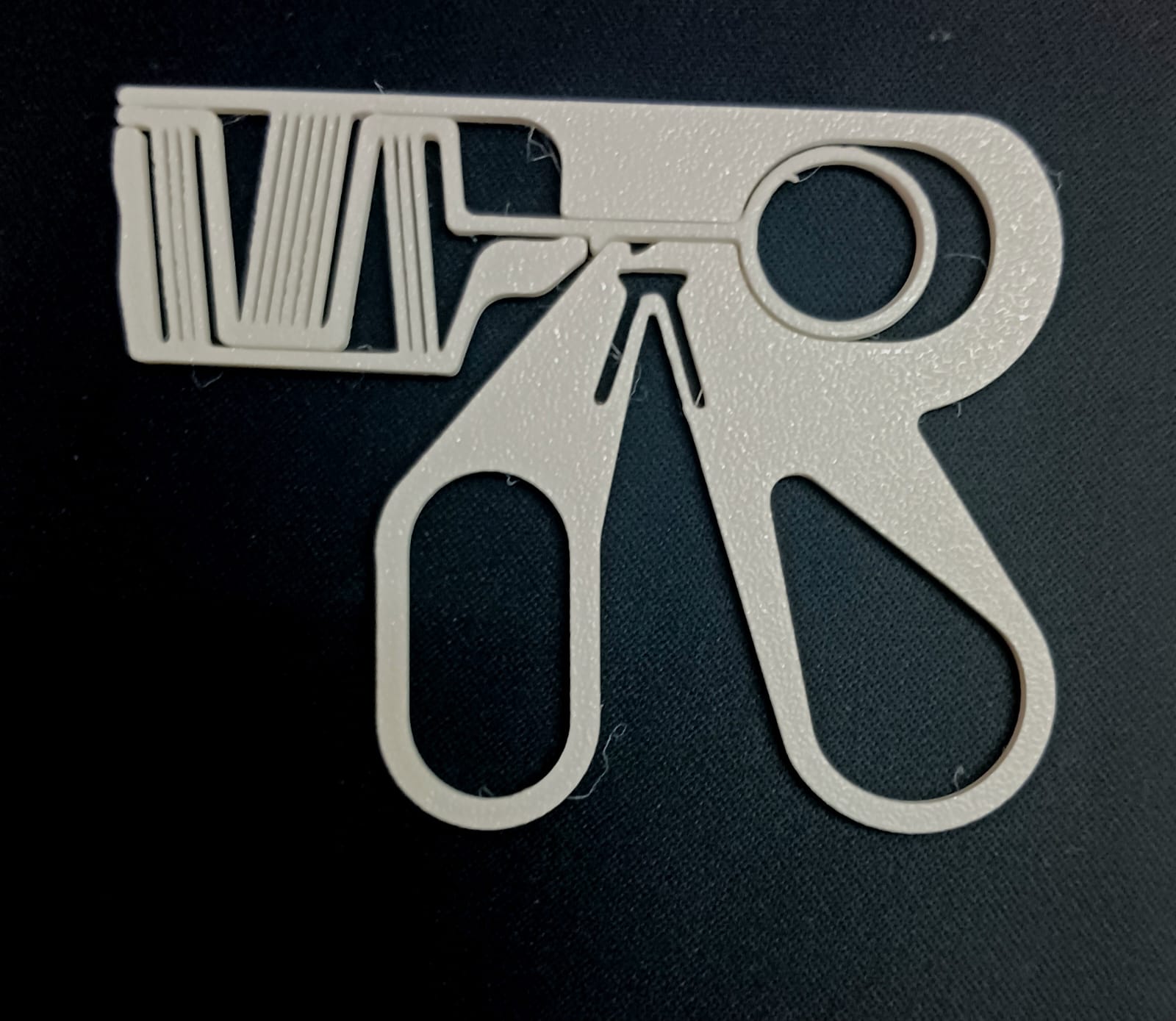}
 		\subcaption{\label{fig:prot-2-rest}}
 	\end{subfigure}
 	\begin{subfigure}{0.3\textwidth}
 		\centering
 		\includegraphics[height=3.7cm,keepaspectratio]{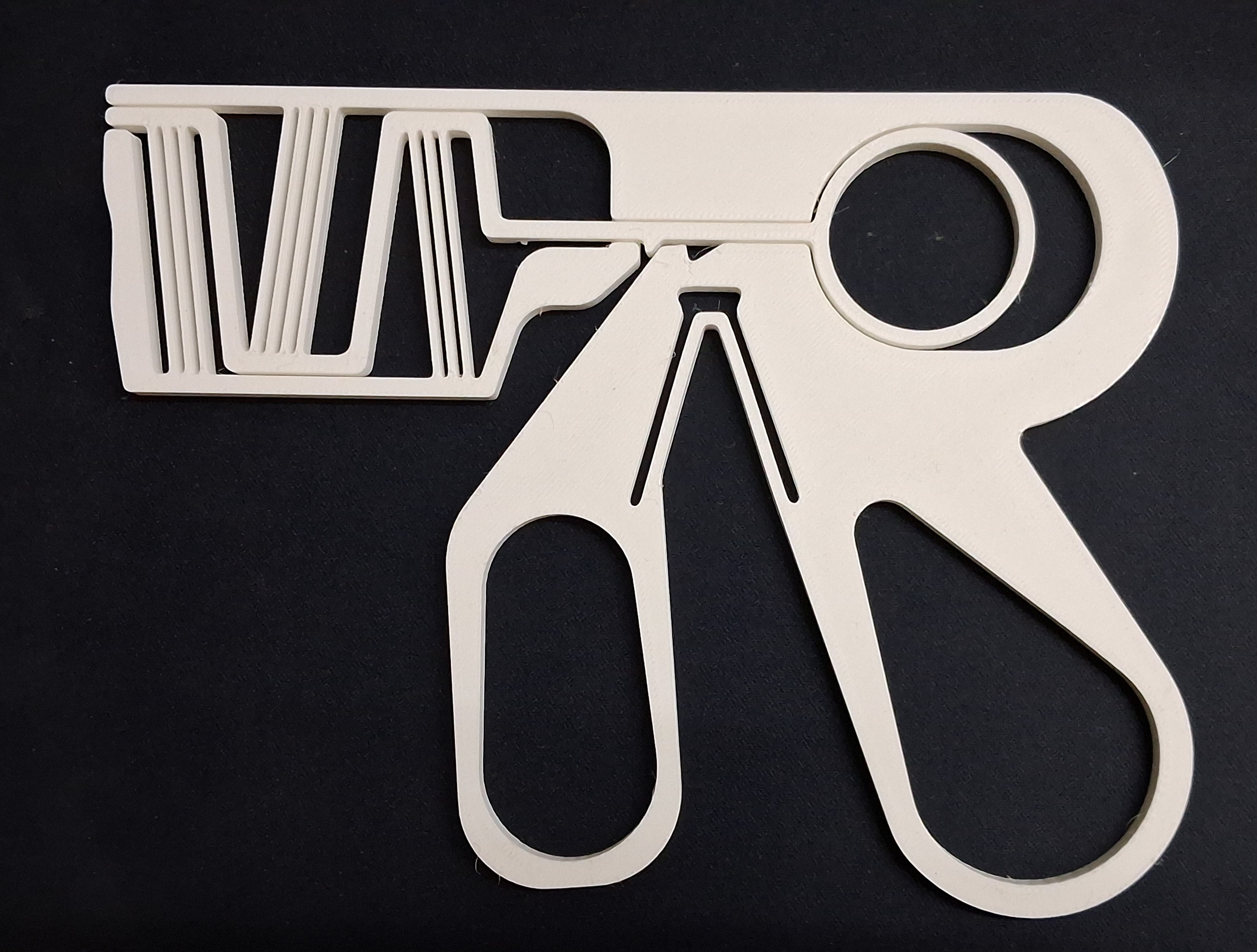}
 		\subcaption{\label{fig:prot-3-rest}}
 	\end{subfigure}
 	\begin{subfigure}{0.3\textwidth}
 		\centering
 		\includegraphics[height=3.7cm,keepaspectratio]{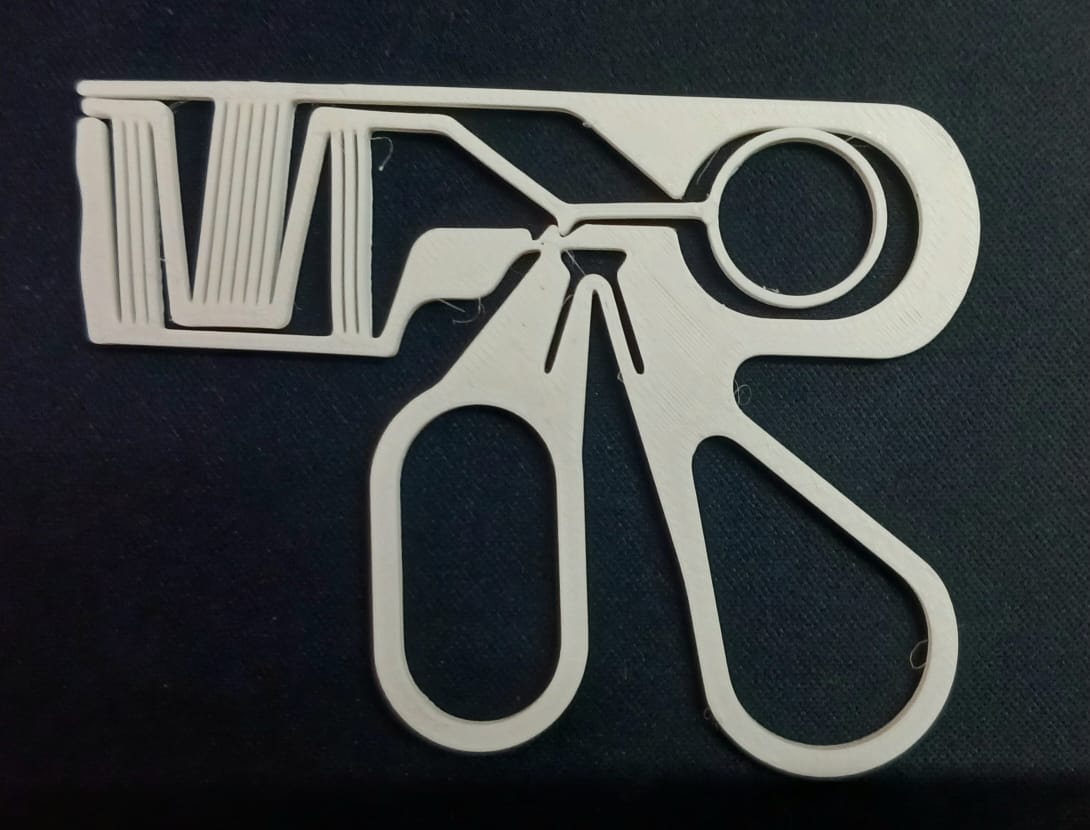}
 		\subcaption{\label{fig:prot-1-press}}
 	\end{subfigure}
 	\begin{subfigure}{0.3\textwidth}
 		\centering
 		\includegraphics[height=3.7cm,keepaspectratio]{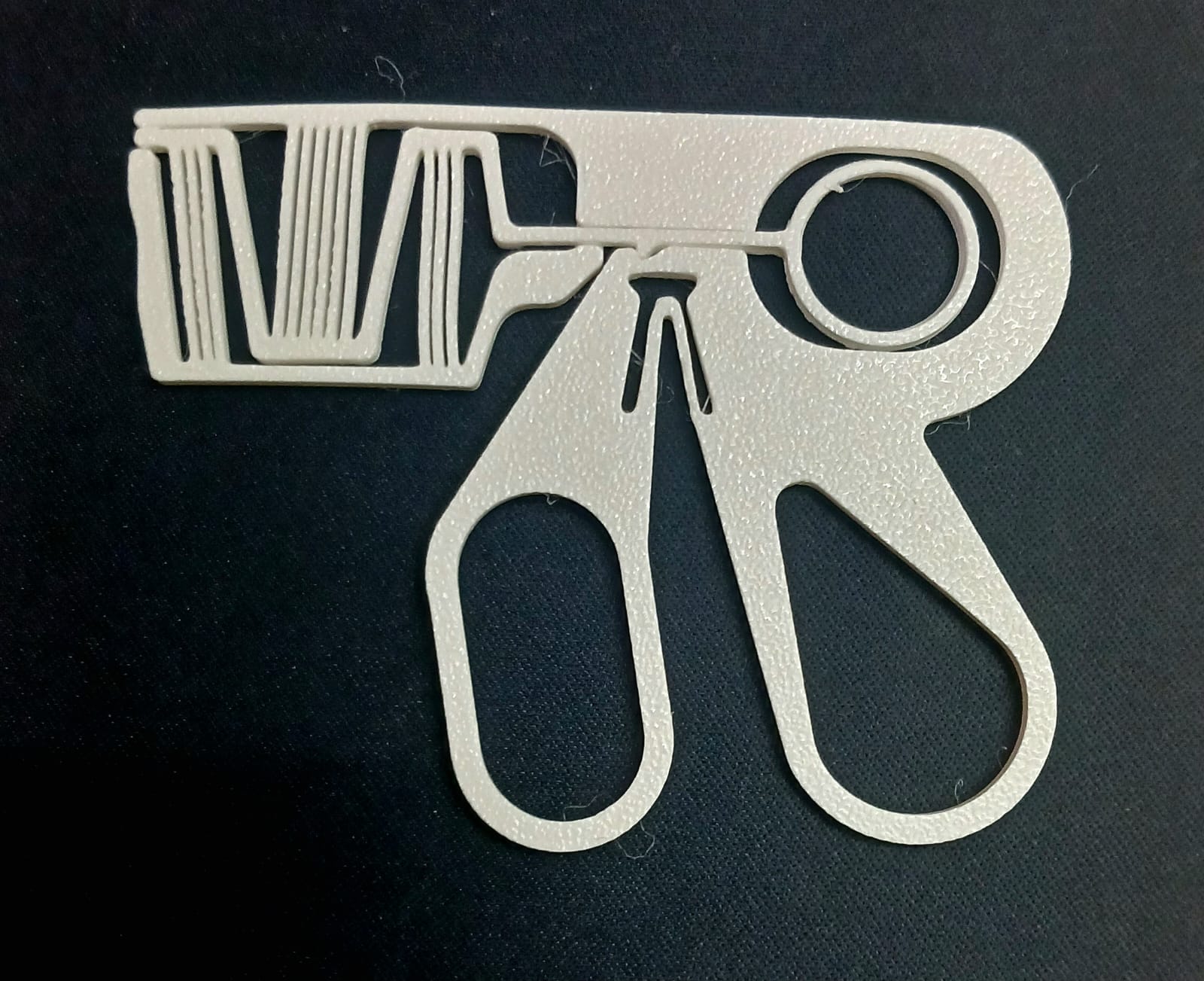}
 		\subcaption{\label{fig:prot-2-press}}
 	\end{subfigure}
 	\begin{subfigure}{0.3\textwidth}
 		\centering
 		\includegraphics[height=3.7cm,keepaspectratio]{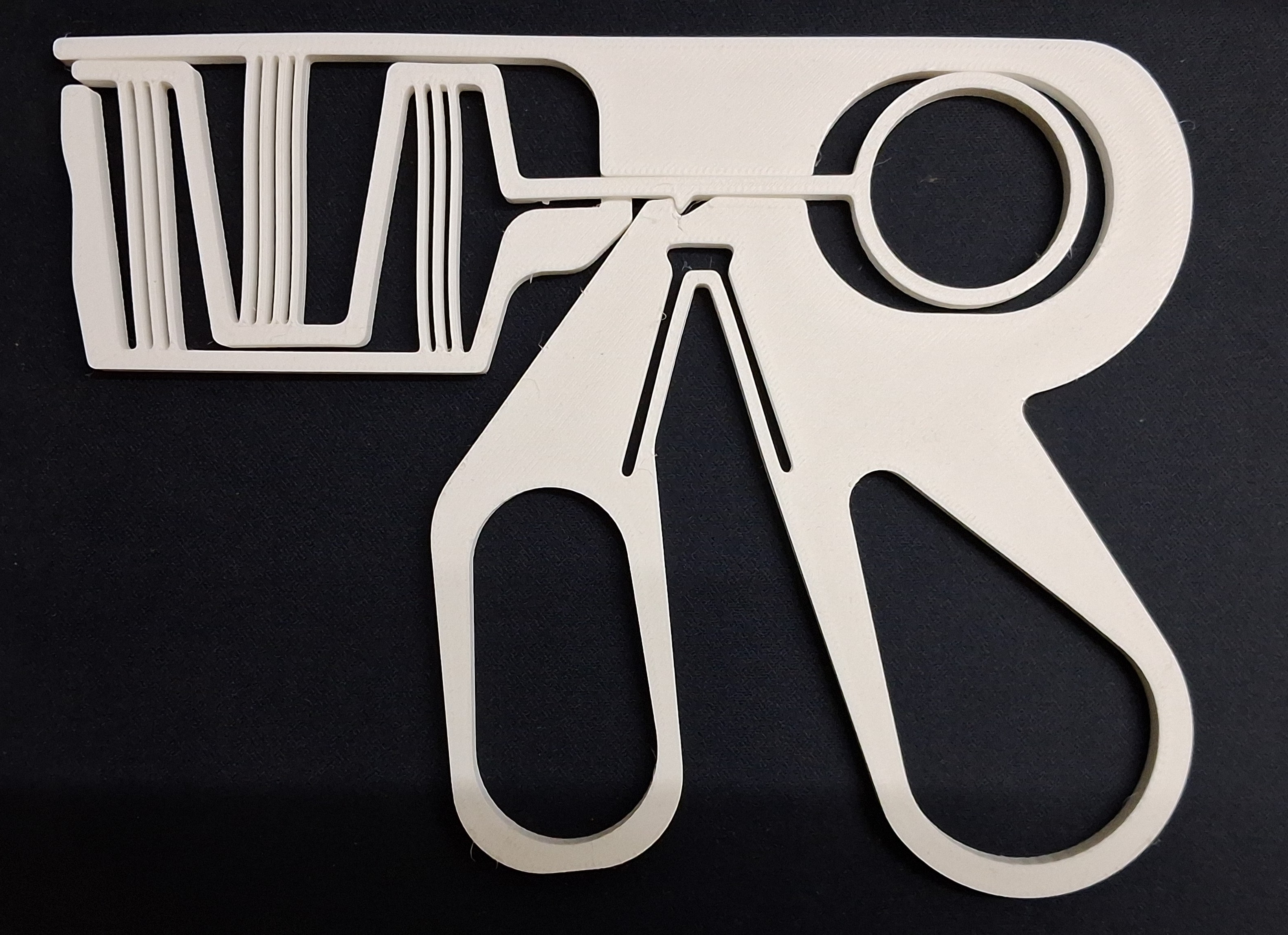}
 		\subcaption{\label{fig:prot-3-press}}
 	\end{subfigure}
 	\caption{The fabricated prototypes in their unstressed states: \ref{fig:prot-1-rest} Prototype 1, \ref{fig:prot-2-rest} Prototype 2 and \ref{fig:prot-3-rest} Prototype 3, and the fabricated prototypes in their stressed states: \ref{fig:prot-1-press} Prototype 1, \ref{fig:prot-2-press} Prototype 2 and \ref{fig:prot-3-press} Prototype 3. The out-of-plane hooks (in Prototype 2 and Prototype 3) permit the trigger to operate satisfactorily. \label{fig:2}}
 \end{figure*}

 The proposed compliant laparoscopic grasper is designed to eliminate the need for the surgeon to apply force continuously during instrument manipulation. The mechanism exhibits two distinct bi-stable configurations: a stressed state and an unstressed state. A circular pull-back ring connected to the control push-rod serves as the primary actuator, enabling the transition of the flexible beams from the unstressed, unbent state to the stressed configuration when pulled proximally (Fig. \ref{fig:1}). In the stressed state, the beams are bent and hence store energy (Figs. \ref{fig:prot-1-press}, \ref{fig:prot-2-press}, \ref{fig:prot-3-press}). Since the beams themselves are not in their bistable configuration, a notch feature is provided, which acts as a latch to prevent the control push-rod from returning the mechanism to its original low-energy state without actuation (Fig.~\ref{fig:1}). 
 
 A trigger mechanism is developed to control the device's return to its unstressed state by releasing the strain energy stored in the beams. When the surgeon pulls the circular ring proximally, the gripper jaws open; conversely, pressing the trigger causes the control push-rod to close the gripper jaws. Therefore, a compliant gripper is employed as the end-effector, wherein jaw opening is achieved by pulling the actuation rod rather than pushing it.  
 
 \par Fused deposition modeling (FDM) is used to fabricate prototypes due to its accessibility, relatively short lead times, cost-effectiveness, and ease of fabricating thin features. Large overhangs require supports during FDM fabrication. Removing these supports for the thin features of our mechanism is difficult, as it may damage the mechanism. Moreover, post-processing time increases. Hence, care is taken to design the mechanism as planar as possible. Typical 3D printers allow for a model size of $256\times256$ mm$^2$. So, the laparoscopic grasper device is designed to fit within a form factor of 200 mm.
 
 \par The development of the compliant laparoscopic grasper is undertaken in two distinct steps. Methodology for the development of the trigger mechanism is provided in Sec.~\ref{subsec:trigger}, while Sec.~\ref{subsec:gripper} explains the design process of the compliant gripper mechanism.
 
 \subsection{Trigger Mechanism} \label{subsec:trigger} \hfill 
 
 The trigger mechanism is responsible for actuating the control rod (Fig.~\ref{fig:1}). It comprises the circular pull-back ring, the bending beams, the trigger handle, and associated parts, such as the notch to prevent un-actuated retraction (Fig.~\ref{fig:1}). The proposed bending beam-based trigger mechanism is inspired by the design presented in Ref.~\cite{robervideo}. Note that the geometry of the flexible beams must be modified to fit within the required form factor and applied force ranges. Since the beams function like parallel guided bistable V-beams~\cite{Liu2021ABC}, we use the Two Elements Beam Constraint (TEBC) Model~\cite{Liu2021ABC} to determine the geometry of the beam for the required actuation force.
 
 \par The TEBC model~\cite{Liu2021ABC} is developed to predict the bistability of symmetric double V-beams, and to determine the actuation forces for a specified beam length, thickness, inclination angle, width, and material. The mathematical equations derived from this model provide the actuation forces required to move the shuttle of a V-beam of a given geometry by a specified travel distance $\Delta Y$. $T$ is the thickness of the beam in the plane, $L$ its length, $W$ its out-of-plane width, and $\theta$ its tilted angle. We define $F_o$, $P_o$, and $M_o$ as the transverse force, axial force, and moment at the guided end, respectively. The related dimensions are normalized, along with the axial deflection $X_o$, the radial deflection $Y_o$, and the thickness~$T$, to convert them into non-dimensional quantities. The corresponding lower-case alphabet of the dimensional quantities represents these non-dimensional quantities. Mathematically, as per~\cite{Liu2021ABC}, one writes:
 
 \begin{equation}
 	\begin{gathered}
 		x_o = \frac{2 X_o}{L} = -\frac{2 \Delta Y \sin{\theta}}{L}, \quad
 		y_o = \frac{2 Y_o}{L} = -\frac{2 \Delta Y \cos{\theta}}{L}, \\  
 		t = \frac{2T}{L}, \quad 
 		f_o = \frac{F_o L^2}{4EI}, \quad
 		p_o = \frac{P_o L^2}{4EI}, \quad
 		m_o = \frac{M_o L}{2EI}
 		\label{eqn1}
 	\end{gathered}
 \end{equation}
 where $E$ is the Young's modulus of elasticity of the material and $I = \frac{WT^3}{12}$ is the moment of inertia of the cross-section. The model also defines an inequality $D_1$ which determines whether this beam of given geometry can be set into its second bistable state by the specified travel distance:
 \begin{equation}
 	D_1 = -4.652 \frac{\cos^2{\theta}}{L^2} \Delta Y + 6.514 \frac{\sin{\theta}}{L} \Delta Y - 21.46 \frac{T^2}{L^2} \geq 0.
 	\label{eqn2}
 \end{equation}
 Note that the satisfaction of $D_1$ is not a sufficient condition for the beam to exist in its second bistable state---it is only a necessary condition. If the required $\Delta Y$ satisfies the inequality $D_1$ (Eq. \ref{eqn2}), then $f_0$ and $p_o$ are calculated as follows:
 \begin{equation}
 	\begin{aligned}
 		f_o &= -4.8618 y_o \\
 		p_o &= -9.8837.
 		\label{eqn3}
 	\end{aligned}
 \end{equation}
 If the inequality $D_1$ (Eq. \ref{eqn2}) is not satisfied, then the following set of equations is used:
 
 \begin{equation}
 	\scalebox{0.75}{$
 		\begin{aligned}
 			\left\{
 			\begin{aligned}
 				k_1 &= \frac{4t^2}{675} + \frac{y_o^2}{42000}\\
 				k_2 &= \frac{16t^2}{45} - \frac{17y_o^2}{2100} - \frac{8x_o}{225}\\
 				k_3 &= \frac{16t^2}{3} - \frac{96y_o^2}{175} - \frac{32x_o}{15}\\
 				k_4 &= -\frac{48y_o^2}{5} - 32x_o
 			\end{aligned}
 			\right.
 			\left\{
 			{\everymath{\textstyle}
 				\begin{aligned}
 					c_1 &= -k_2^2 + 3k_1k_3\\
 					c_2 &= \Biggl[ -2k_2^3 + 9k_1k_2k_3 - 27k_1^2k_4 +\\
 					&\quad\sqrt{4c_1^3 + \left( -2k_2^3 + 9k_1k_2k_3 - 27k_1^2k_4 \right) ^2}\Biggr] ^{1/3}\\
 					p_1 &= -\frac{k_2}{3k_1}
 					+ \frac{\sqrt[3]{2c_1}}{3k_1c_2}
 					+ \frac{c_2}{3\sqrt[3]{2k_1}}
 				\end{aligned}
 			}
 			\right.
 		\end{aligned}
 		$}
 	\label{eqn4}
 \end{equation}
 
 \begin{equation}
 	\scalebox{0.8}{$
 		\begin{aligned}
 			f_o &= \dfrac{y_o}{2} \left[\left(12 + \dfrac{6}{5} p_1 + \dfrac{1}{700}p_1 ^2\right) - \dfrac{\left(-6 - \dfrac{1}{10}p_1 + \dfrac{1}{1400}p_1 ^2\right)^2}{\left(4+ \dfrac{2}{15}p_1 - \dfrac{11}{6300}p_1^2\right)} \right]\\
 			p_o &= -\dfrac{k_2}{3k_1} - \dfrac{\sqrt[3]{2}c_1}{3 k_1 c_2} + \dfrac{c_2}{3 \sqrt[3]{2 k_1}}.
 			\label{eqn5}
 		\end{aligned}
 		$}
 \end{equation}
 The force $F$ required to displace the bistable beam is thus defined as~\cite{Liu2021ABC}:
 \begin{equation}
 	F = - \left(\frac{4 E I f_o}{L^2} \cos{\theta} + \frac{4 E I p_o}{L^2} \sin{\theta} \right)
 	\label{eqn6}
 \end{equation}
 
 \begin{figure*}[!t]
 	\centering
 	\begin{subfigure}{0.495\textwidth}
 		\centering    \includegraphics[width=\linewidth,height=0.24\textheight,keepaspectratio]{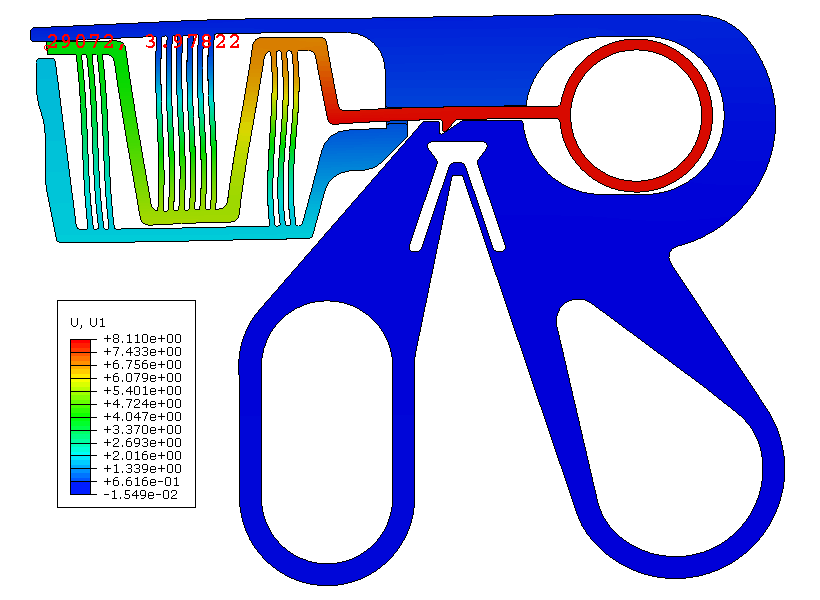}
 		\subcaption{\label{fig:12beams}}        
 	\end{subfigure}
 	\begin{subfigure}{0.495\textwidth}
 		\centering
 		\includegraphics[width=\linewidth,height=0.24\textheight,keepaspectratio]{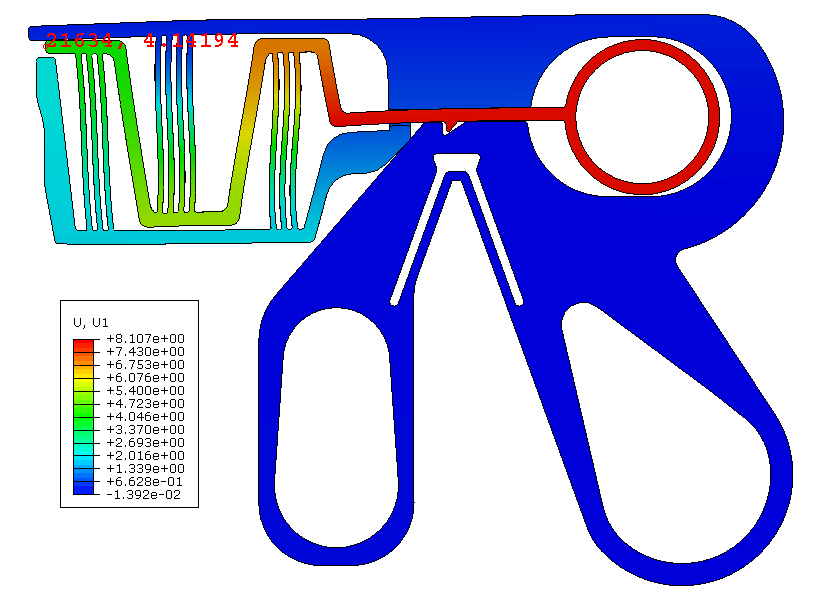}
 		\subcaption{\label{fig:10beams}}        
 	\end{subfigure}
 	\caption{The displacement profiles of Prototype~2 (Fig.~\ref{fig:12beams}) and Prototype~3 (Fig.~\ref{fig:10beams}) along the $x$ direction (the control push-rod direction) in its stressed state. The displacement of the shuttle in this direction is indicated as the second value in the annotation in the plots. The shuttle displacement is higher for Prototype~3 (Fig.~\ref{fig:10beams}) compared to Prototype~2 (Fig.~\ref{fig:12beams}).}
 \end{figure*}
 
 \par In our application, single V-beams are used, which we call I-beams. Moreover, the I-beams must not be set into their second bistable state, as this would significantly reduce the energy stored in the bent beams and cause the mechanism to fail to return to its original state once actuated by the trigger. Hence, the TEBC Model~\cite{Liu2021ABC} is used purely for the estimation of an initial geometry corresponding to a desired travel distance of 5 mm and an approximate order of magnitude of force of 20 N~\cite{olig2023output, van2012indirect}. Since the mathematical formulations in the TEBC Model~\cite{Liu2021ABC} correspond to double V-beams, a simple approximation of halving the output force (by invoking symmetry) is taken for this case of single I-beams, as well as multiplying this halved force by the number of beams to obtain the total force applied on the circular pull-back ring of the device by the surgeon. An initial configuration is chosen for 12 bending beams. The geometric parameters obtained using the TEBC Model formulation are indicated as follows: the length is 40~mm, thickness 1.2~mm, width 5~mm, tilt angle 7\textdegree, travel distance 5~mm, and force 21.6~N.
 
 \par The circular pull-back ring, trigger handle, and latching notch are incorporated to accommodate the present dimensional specifications. The trigger is designed with thin features that act as flexures, allowing the trigger handle to flex. While finite element analyses are used to conduct a preliminary study of the circular pull-back ring's behavior and to compare shuttle displacements across different prototypes, the functionality of the trigger mechanism is determined through iterative prototyping. This is because accurate finite element analyses need to capture the exact snap-through position of the notch feature, as well as the high-strain-rate deformation during the release of strain energy upon trigger actuation.
 \subsubsection{3D-printing steps and comparative study \label{subsubsec:3dprint}}
 We present a hybrid analytical-experimental methodology to design the proposed TiBCLaG. The TEBC Model~\cite{Liu2021ABC} provides the initial beam geometry, but several design parameters, such as notch geometry, trigger kinematics, and multi-beam displacement sharing, involve snap-through and nonlinear contact characteristics that are challenging to predict analytically or computationally at early design stages. Therefore, we use physical prototyping to predict these parameters. Design variables are modified between iterations, and the resulting change in behavior is used to establish a design rule for the next iteration.
 
 The 3D computer-aided design (CAD) model is prepared using Autodesk Fusion and fabricated by fused deposition modeling in a Bambu Lab P1S using polylactic acid (PLA) filament with a nozzle temperature of 220 \textdegree C, bed temperature of 55 \textdegree C, print speed of 400 mm/s, layer height 0.2 mm, and 15\% infill density. The model is scaled by half because the objective is to validate the functioning mechanism and not quantify the numerical values of load or deformation.
 
 \begin{figure}[h!]
 	\centering\includegraphics[width=1\linewidth]{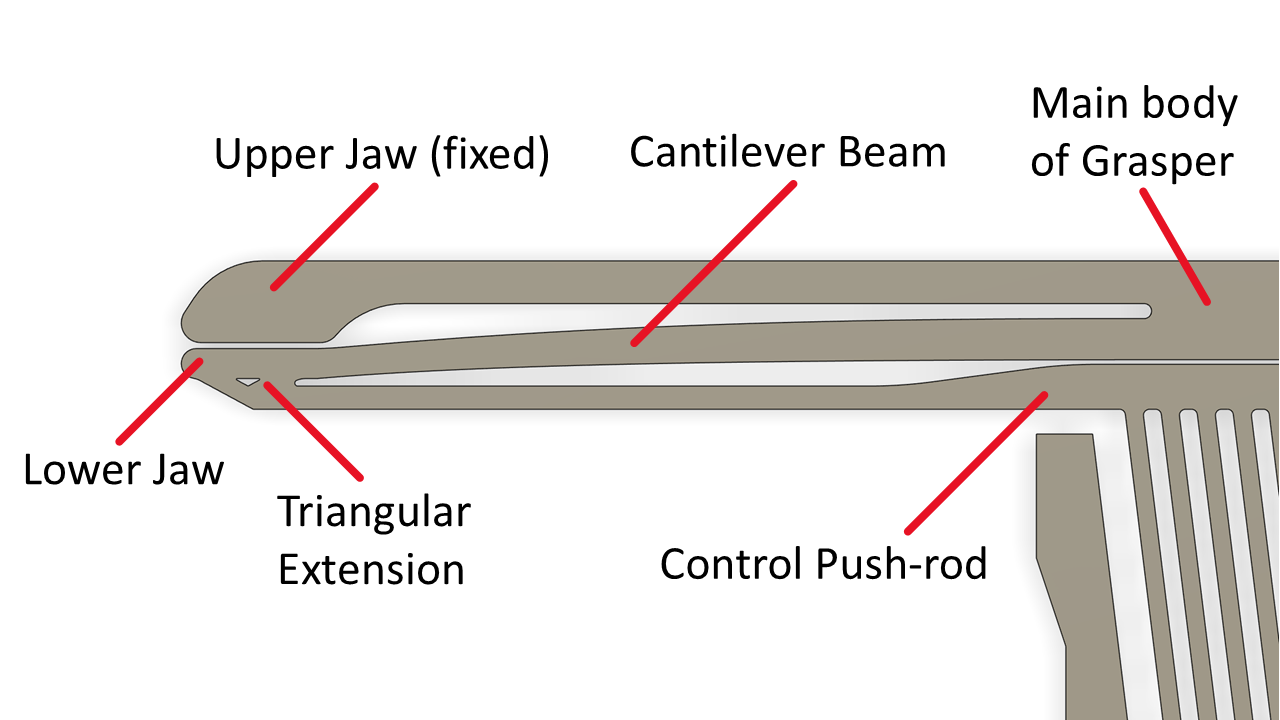}
 	\caption{Plan view of the 3D CAD model of the Gripper.\label{fig:4}}
 \end{figure}
 
 \begin{figure*}[h!]
 	\centering
 	\begin{subfigure}{0.495\textwidth}
 		\centering
 		\includegraphics[width=\linewidth,height=0.24\textheight,keepaspectratio]{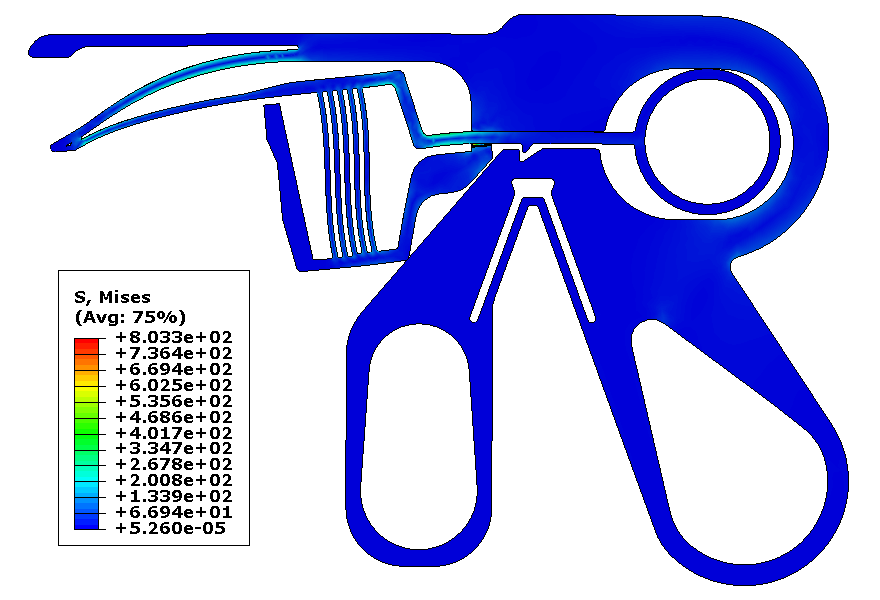}
 		\subcaption{\label{fig:5beams}}
 	\end{subfigure}\hfill
 	\begin{subfigure}{0.495\textwidth}
 		\centering
 		\includegraphics[width=\linewidth,height=0.24\textheight,keepaspectratio]{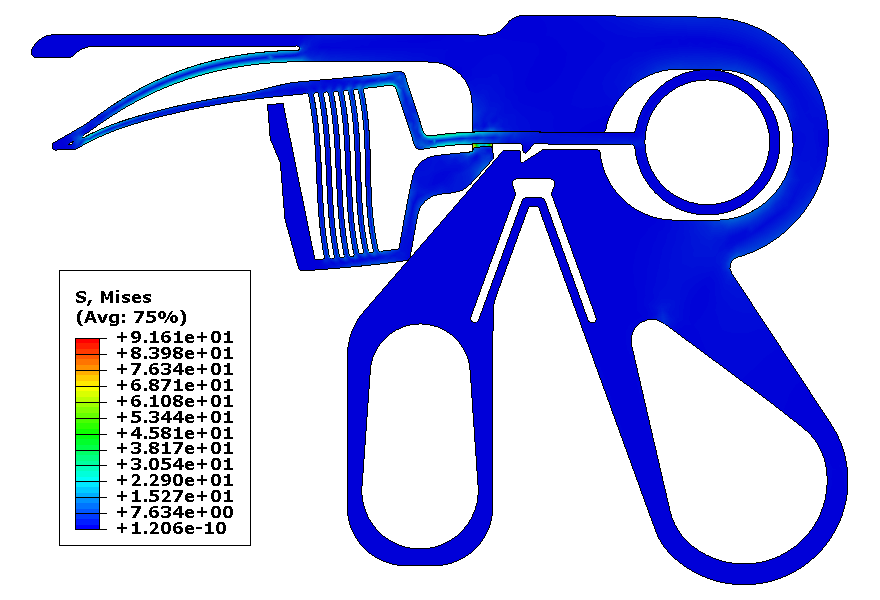}
 		\subcaption{\label{fig:6beams}}
 	\end{subfigure}
 	\caption{The FEM analysis results of the five-beam and the final six-beam configuration. The stresses induced in the five-beam configuration (Fig. \ref{fig:5beams}) are much higher than in the six-beam configuration (Fig. \ref{fig:6beams}). The six-beam configuration (Fig. \ref{fig:6beams}) presents a suitable compromise between the displacement of the gripper jaws and the stresses induced.\label{fig:5v6BeamComparison}}
 \end{figure*}
 \begin{figure*}[h!]
 	\centering
 	\begin{subfigure}{0.495\textwidth}
 		\centering
 		\includegraphics[width=\linewidth,height=0.24\textheight,keepaspectratio]{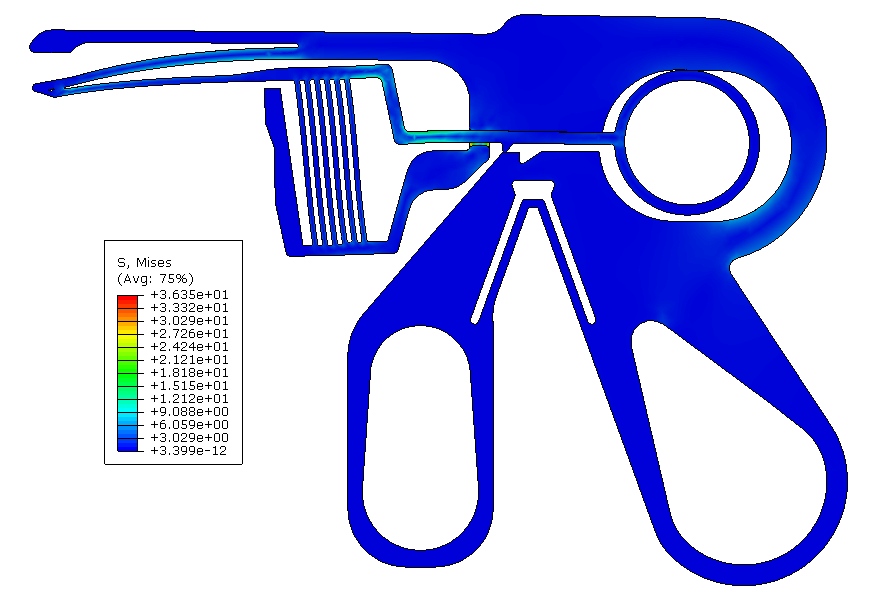}
 		\subcaption{\label{fig:load32}}
 	\end{subfigure}\hfill
 	\begin{subfigure}{0.495\textwidth}
 		\centering
 		\includegraphics[width=\linewidth,height=0.24\textheight,keepaspectratio]{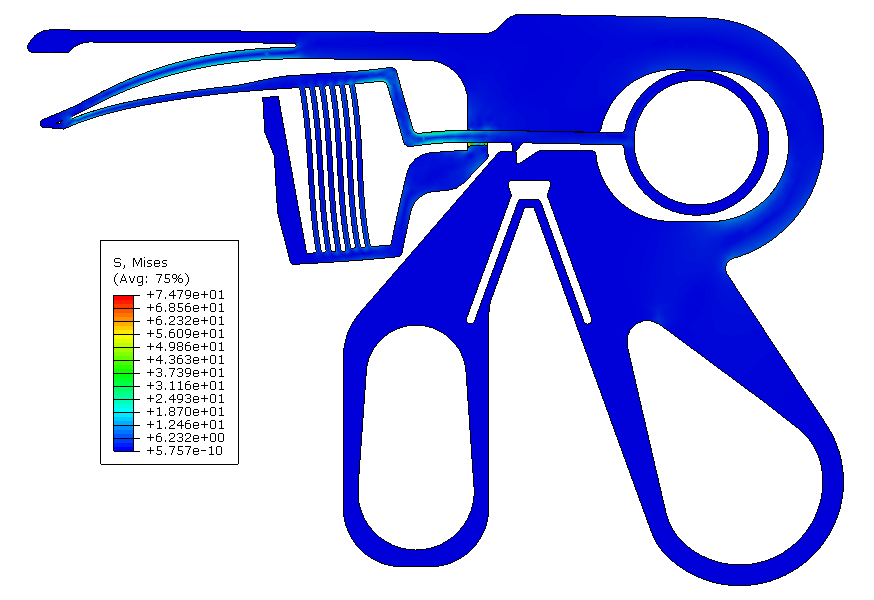}
 		\subcaption{\label{fig:load64}}
 	\end{subfigure}
 	\caption{Snapshots of the FEM analysis of the final laparoscopic grasper prototype at the end of \ref{fig:load32} Step 1 (3.2mm displacement) and \ref{fig:load64} Step 2 (6.4mm displacement). \label{fig:midloads}}
 \end{figure*}
 \par The printed physical Prototype 1 (Fig.~\ref{fig:prot-1-rest} and Fig.~\ref{fig:prot-1-press}) provides valuable information on the actuation of the notch feature. Ideally, when pressed, the trigger must lower the notch so that the shuttle can cross over to the low-energy state. However, with this design, the actuation of the trigger pushes the notch upward, further blocking the shuttle. Hence, the beams do not return to the original low-energy state when the trigger is actuated. Thus, we redesign the model by altering the geometry around the trigger, Prototype 2 (Fig.~\ref{fig:prot-2-rest} and Fig.~\ref{fig:prot-2-press}). An out-of-plane hook is added to connect the flexible beam geometry to the circular pull-back ring housing, permitting the trigger to move freely. It is observed that actuating the trigger results in an unintuitive motion of the rear handle. The front handle should behave like a trigger, moving when pressed, and the rear handle serves purely as a support. So, we add more material to the joint between the rear handle and the pull-back ring housing to increase its bending stiffness. More material is also added to the control push-rod housing to constrain its transverse motion. The CAD model is once again scaled down by half and fabricated using the same settings and material.
 
 \par The TEBC Model~\cite{Liu2021ABC}, as mentioned previously, is developed for the double V-beam mechanism. It does not account for the effects of multiple beams. This is clearly demonstrated in the physical Prototype 2~ (Figs. \ref{fig:prot-2-rest} and \ref{fig:prot-2-press}) and in the analysis result in Fig. \ref{fig:12beams}, performed using the finite element framework discussed in detail in Sec.~\ref{subsubsec:fem}. Due to the additional beams, the displacement of the control push-rod at the circular pull-back ring is not transferred to the shuttle end itself; i.e., the shuttle does not move as much as the circular pull-back ring. Since the shuttle's travel distance is a critical actuation for the compliant gripper  (Sec.~\ref{subsec:gripper}), it is necessary to increase the mechanism's compliance. Hence, the next prototype, Prototype 3,  is fabricated in PLA with only 10 beams instead of 12 to observe the difference (Figs. \ref{fig:prot-3-rest} and \ref{fig:prot-3-press}). The same settings as before are used.
 
 \par This prototype clearly indicates that decreasing the number of beams increased the effective displacement of the shuttle (Fig. \ref{fig:10beams}). Overall, the iterative prototyping process and corresponding analysis results reveal that although the TEBCM provides a valuable baseline for initial geometry selection, it does not fully capture the collective stiffness and displacement-sharing effects arising from multiple interacting beams. Consequently, the beam count is reduced to achieve the required shuttle travel, ensuring reliable actuation of the compliant gripper described in Sec.~\ref{subsec:gripper}. These results underscore the necessity of experimental validation when extending analytical models beyond their original assumptions.

 \begin{table*}[h!]
 	\centering
 	\caption{Stresses and Displacements during Retraction of Trigger\label{tab:2}}
 	\begin{tabular}{c c c}
 		\toprule
 		Trigger Displacement [mm] & Maximum Stress [MPa] & Jaw Displacement [mm] \\
 		\midrule
 		3.2 & 36.35 & 7.13 \\
 		6.4 & 74.79 & 15.99 \\
 		8 & 91.61 & 20.52 \\
 		\bottomrule
 	\end{tabular}
 \end{table*}

 \begin{figure*}[h!]
 	\centering
 	\begin{subfigure}{0.495\textwidth}
 		\centering
 		\includegraphics[width=\linewidth,height=0.24\textheight,keepaspectratio]{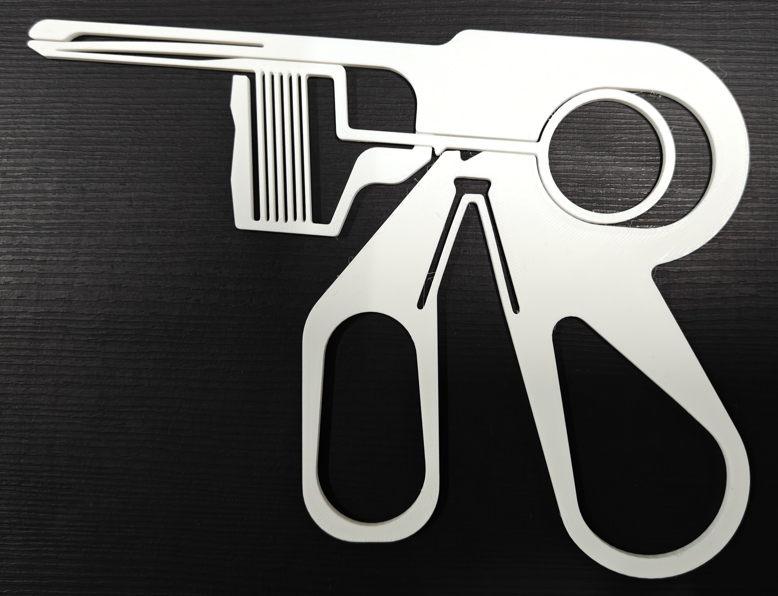}
 		\subcaption{\label{fig:finalrest}}
 	\end{subfigure}\hfill
 	\begin{subfigure}{0.495\textwidth}
 		\centering
 		\includegraphics[width=\linewidth,height=0.24\textheight,keepaspectratio]{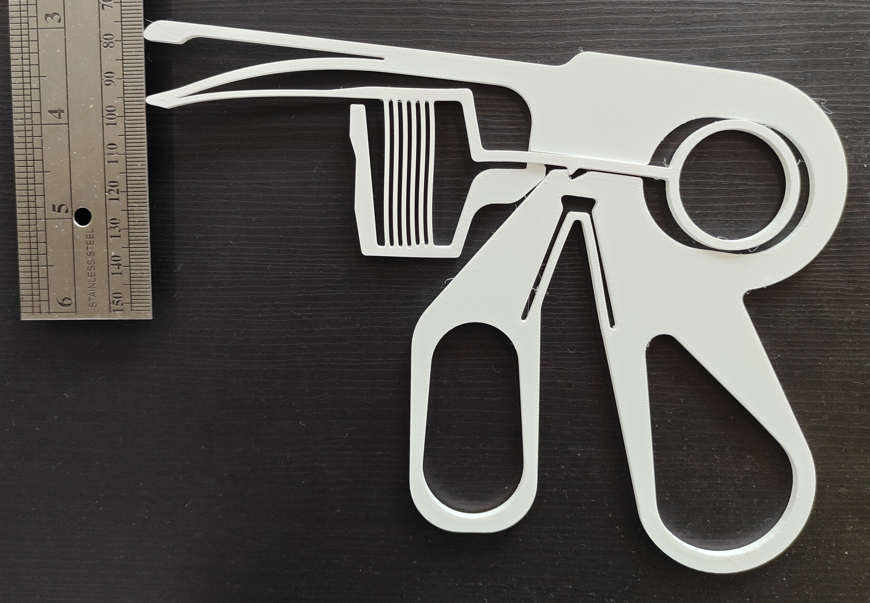}
 		\subcaption{\label{fig:finalstress}}
 	\end{subfigure}
 	\begin{subfigure}{0.95\textwidth}
 		\centering
 		\begin{tikzpicture}[scale=0.95]
 			\begin{axis}[
 				xlabel={\large Ring displacement (mm)},
 				ylabel={\large Jaw displacement (mm)},
 				xmin=0,xmax=10,
 				grid=major,
 				legend style={at={(0.2,0.8)},anchor=north}
 				]
 				\addplot[smooth,{black}, line width=1.5pt, mark = star] coordinates {
 					(1,1.94451)
 					(2,3.10728)
 					(3,4.63333)
 					(3.8,6.93566)
 					(4.6,9.26203)
 					(5.4,11.5909)
 					(6.2,13.8994)
 					(7,16.4628)
 					(7.5,19.5011)
 					(8,20.52)}; \addlegendentry{\large FEM}
 				\addplot[smooth,{blue}, line width=1.5pt, mark = square] coordinates {
 					(1,2)
 					(2,3)
 					(3,5)
 					(4,6)
 					(5,8)
 					(6,10)
 					(7,12)
 					(8,15)}; \addlegendentry{\large Experiment}
 			\end{axis}
 		\end{tikzpicture}
 		\subcaption{\label{fig:comparison}}
 	\end{subfigure}
 	\caption{Results obtained from the final printed prototype. The final prototype of the laparoscopic grasper is presented in its two states, \ref{fig:finalrest} closed and \ref{fig:finalstress} open. Comparison of the gripper jaw displacement with the circular pull-back ring displacement obtained from finite element analysis and experiments is presented in Fig.~\ref{fig:comparison}. \label{fig:8-finalprototypes}}
 \end{figure*}
 
 \begin{table*}[t!]
 	\centering
 	\caption{Pictorial Comparison between the FEM and experimental deformed profiles of the laparoscopic grasper. The comparison between the displacements of the jaw and the ring are illustrated in Fig.~\ref{fig:comparison}}.
 	\begin{tabular}{|c|cc|}
 		\hline
 		\multirow{2}{*}{Displacement [mm]} & \multicolumn{2}{c|}{Deformed Profiles} \\ \cline{2-3}
 		& \multicolumn{1}{c|}{FEM} & Experimental                                 \\ \hline
 		1                                      & \multicolumn{1}{c|}{\includegraphics[scale=0.15]{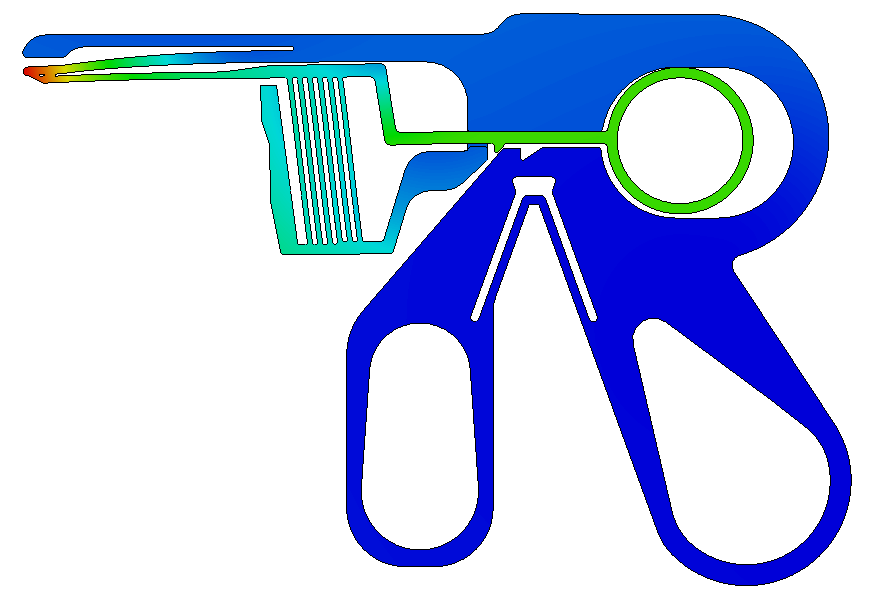}}             &   \includegraphics[angle=90, scale=0.023]{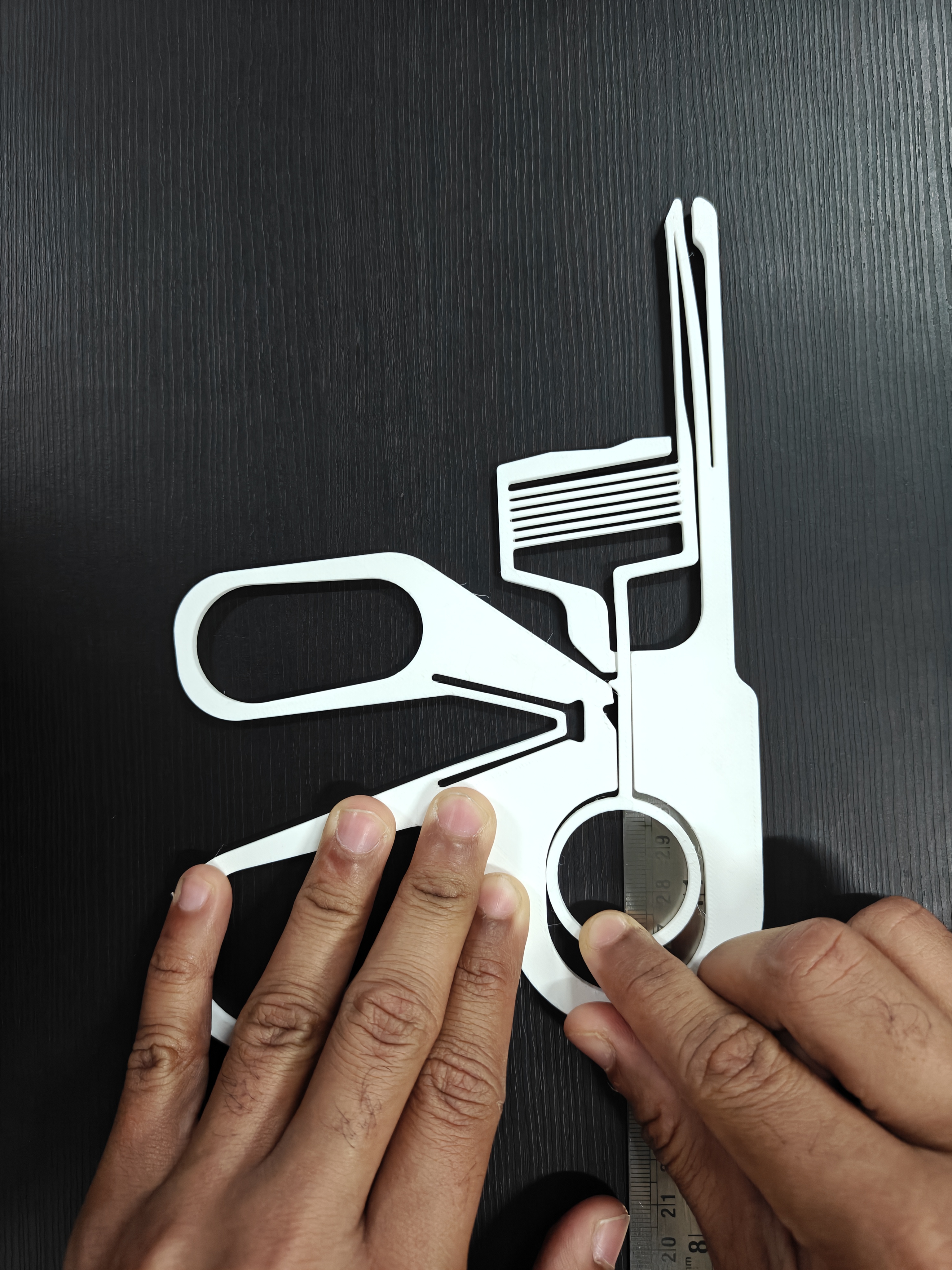}\\ \hline
 	2                                      & \multicolumn{1}{c|}{\includegraphics[scale=0.15]{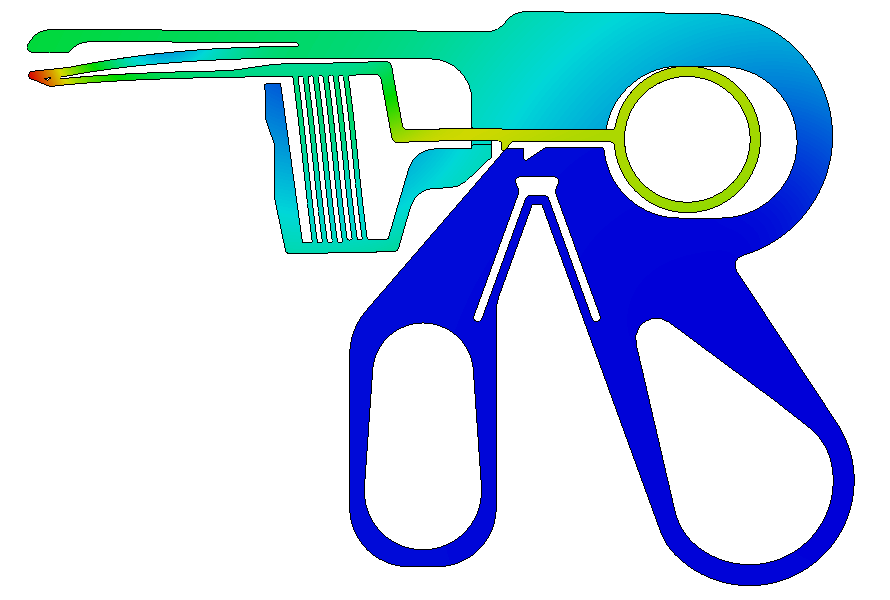}}             &    \includegraphics[angle=90, scale=0.023]{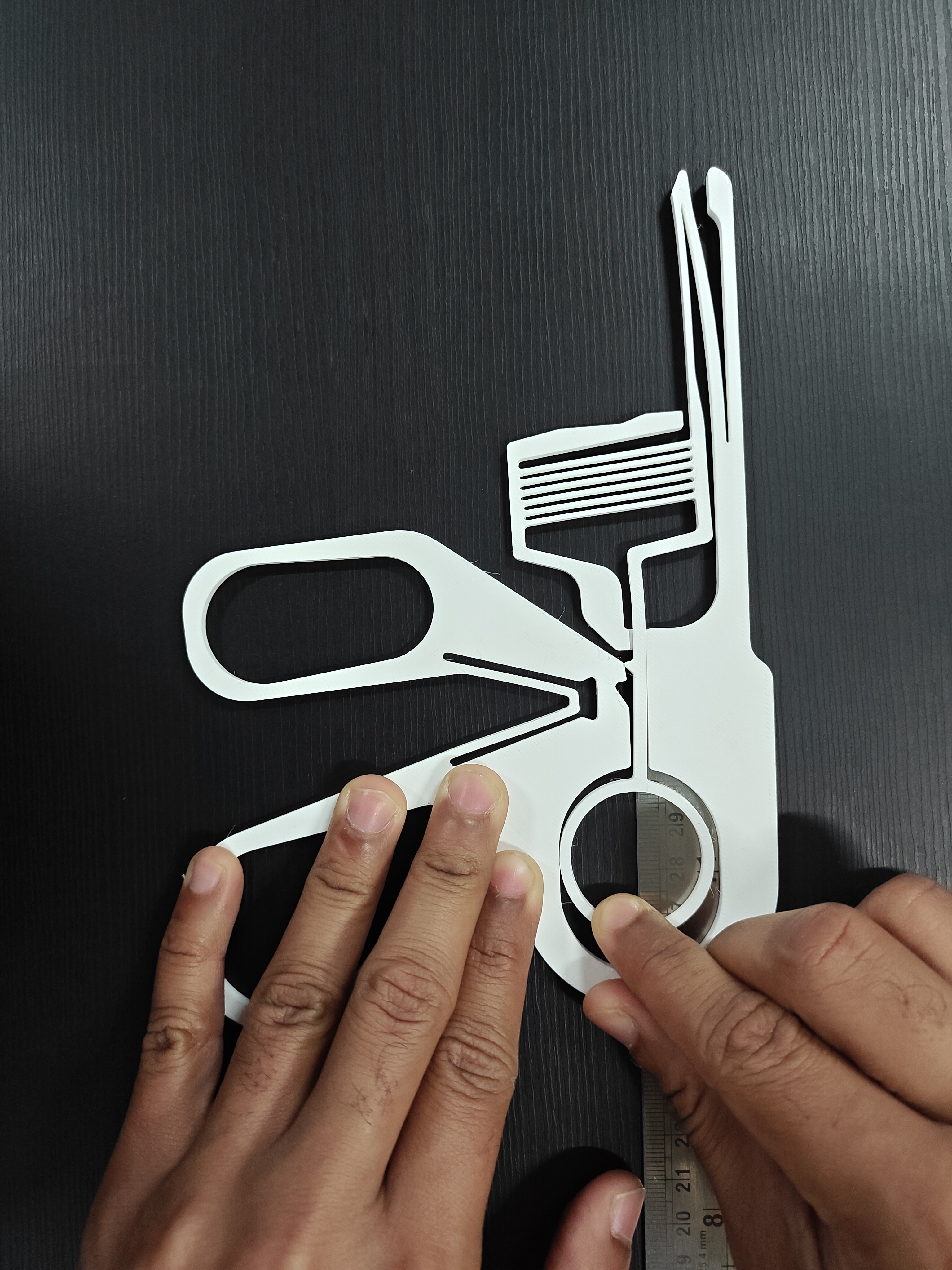}\\ \hline
 	3                                      & \multicolumn{1}{c|}{\includegraphics[scale=0.15]{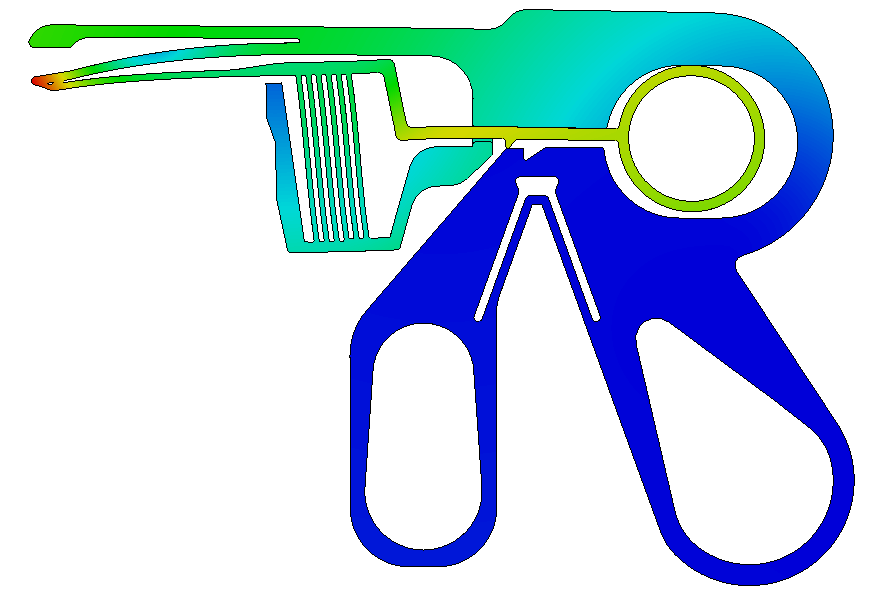}}             &    \includegraphics[angle=90, scale=0.023]{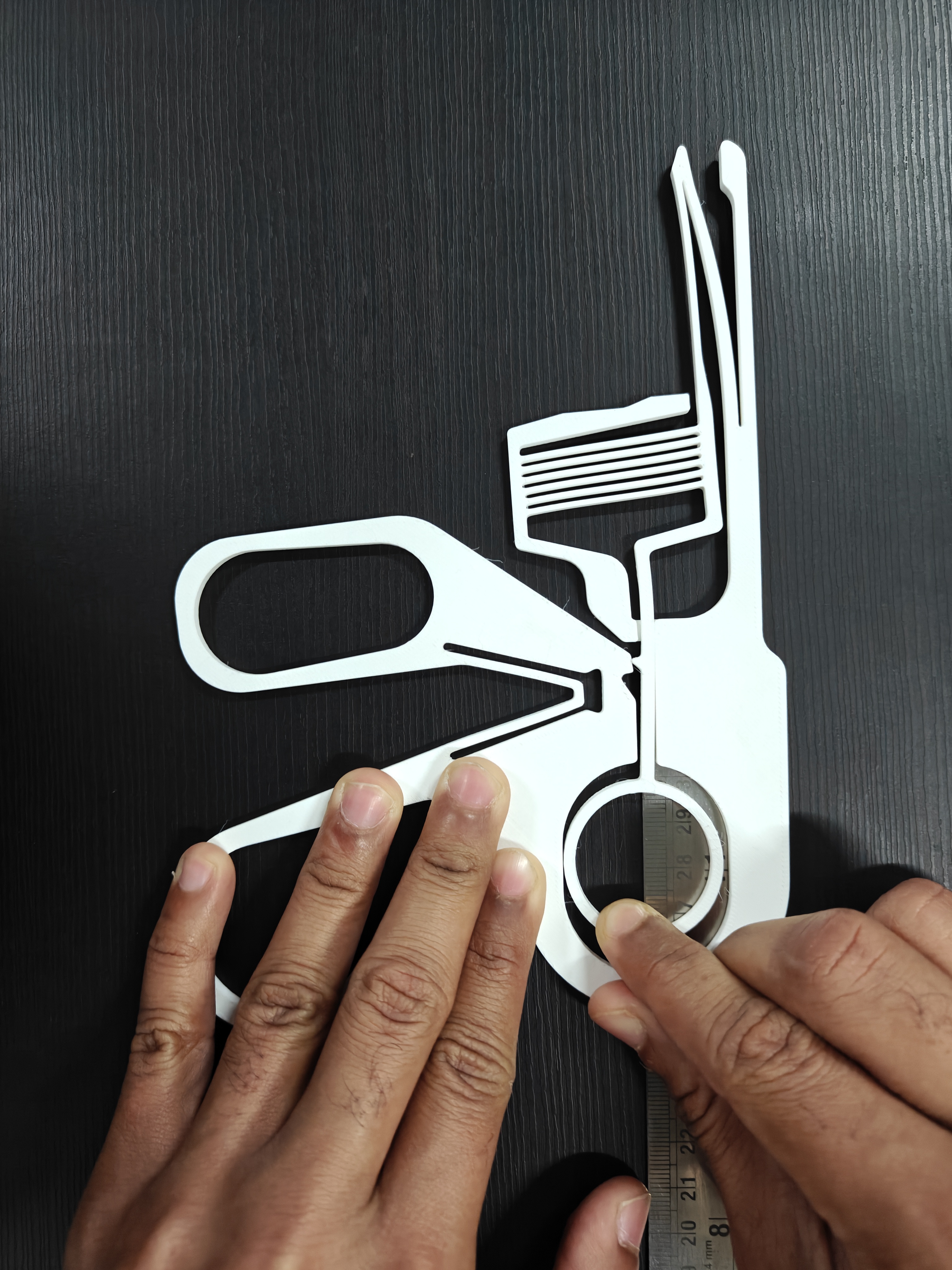}\\ \hline
 	4                                      & \multicolumn{1}{c|}{\includegraphics[scale=0.15]{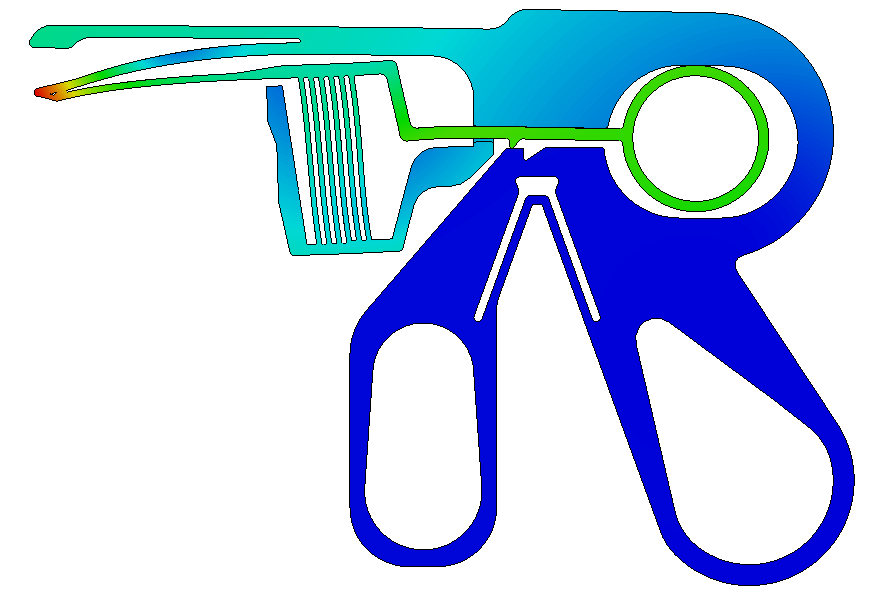}}             &    \includegraphics[angle=90, scale=0.023]{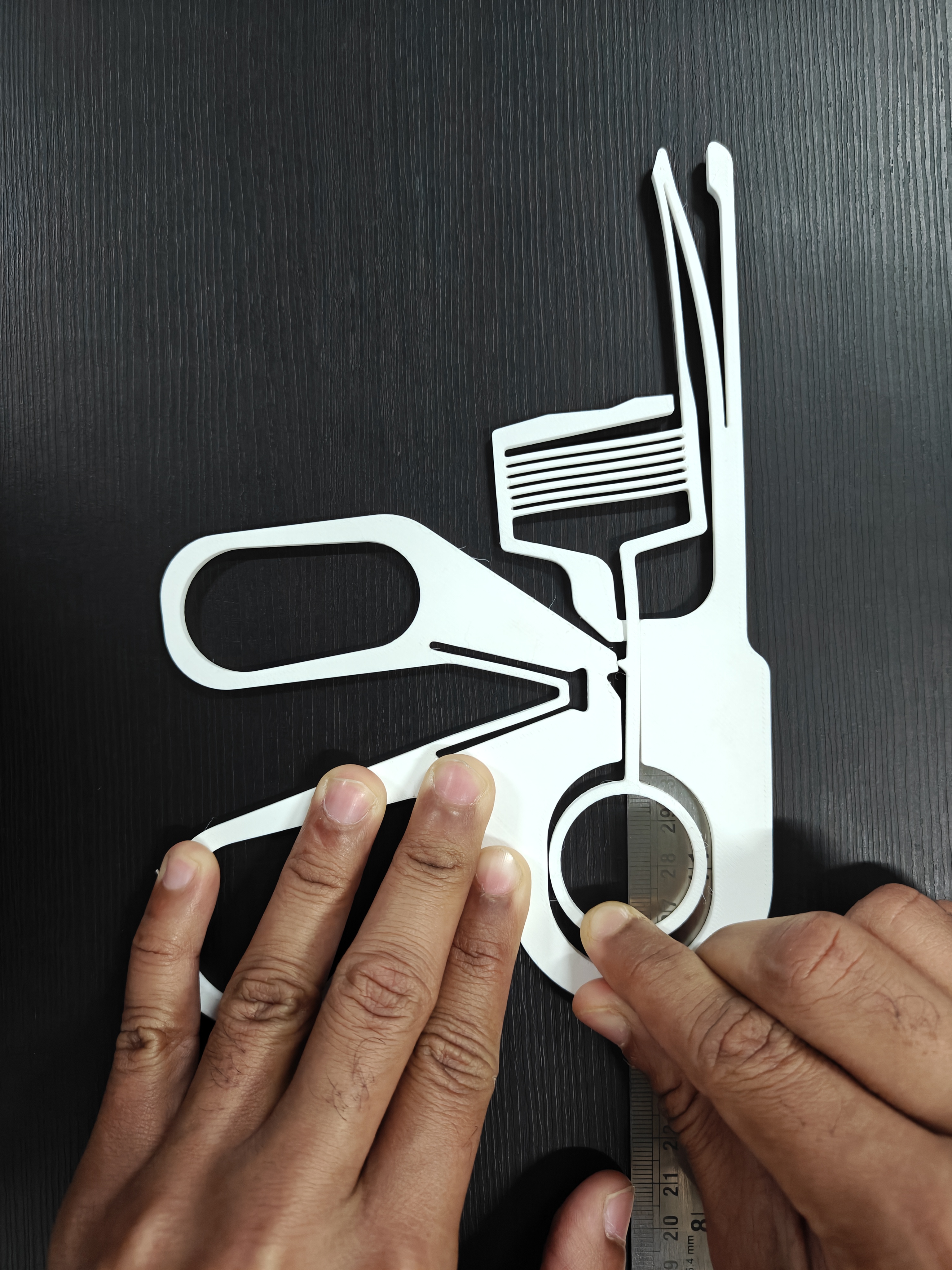}\\\hline
 	5                                      & \multicolumn{1}{c|}{\includegraphics[scale=0.15]{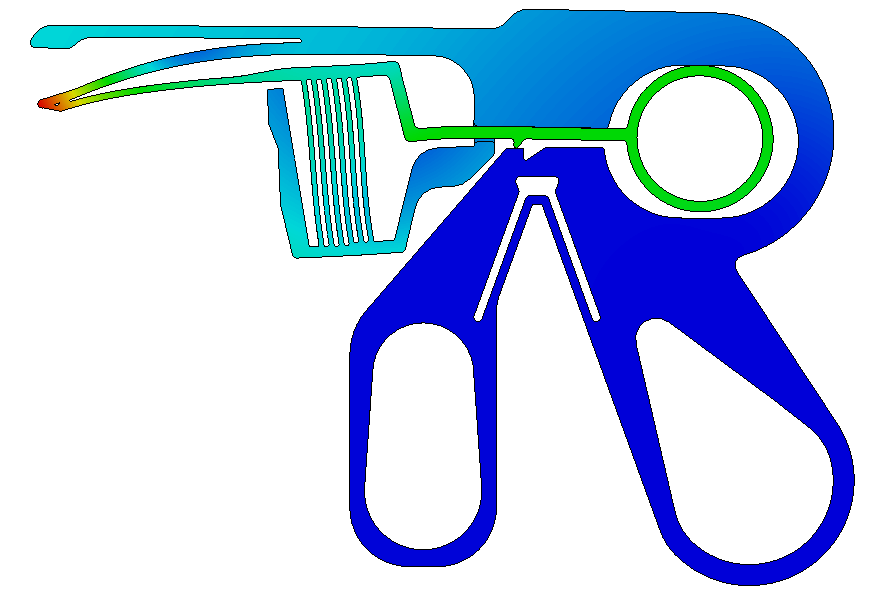}}             &     \includegraphics[angle=90, scale=0.023]{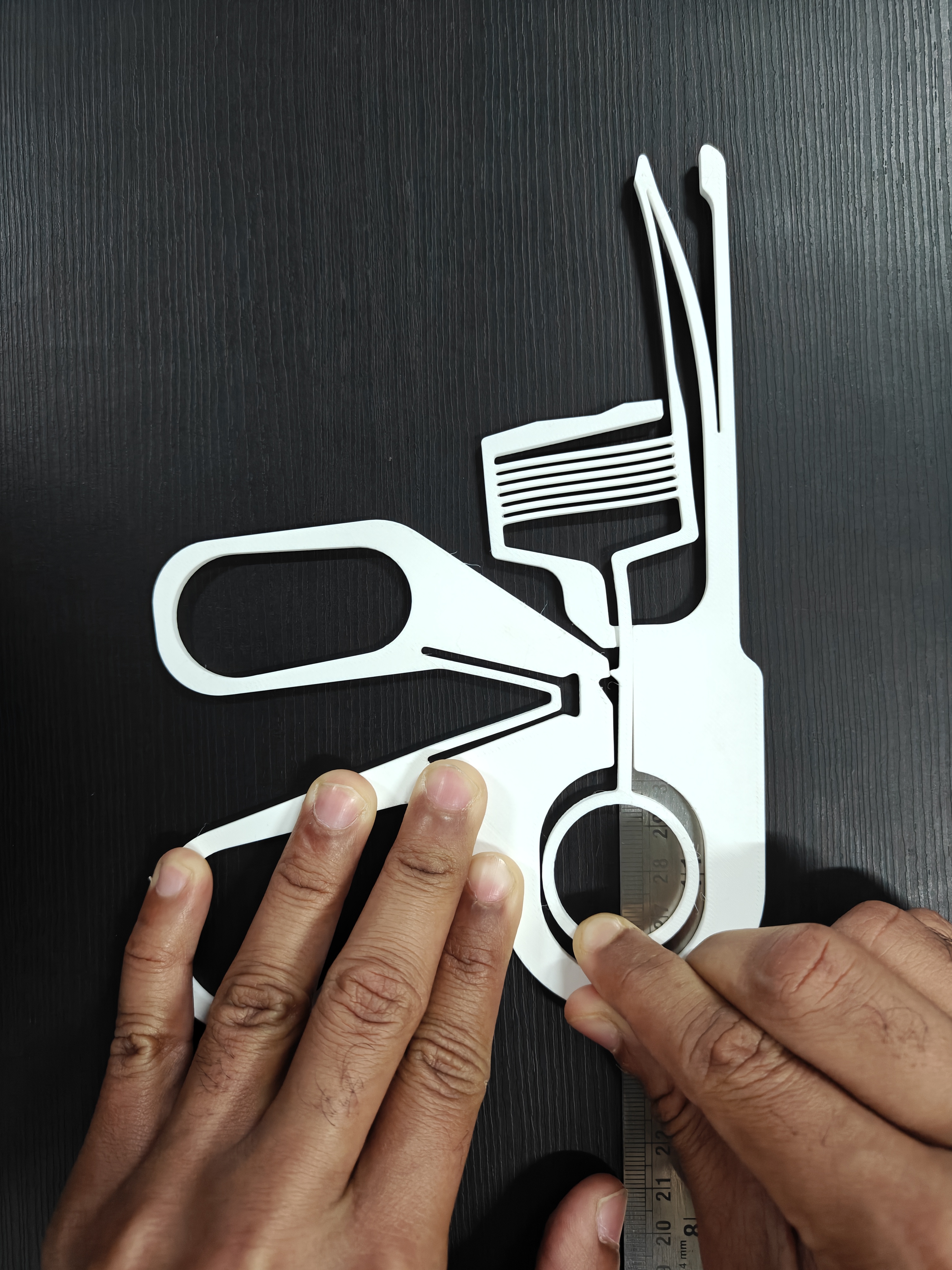}\\\hline
 	6                                      & \multicolumn{1}{c|}{\includegraphics[scale=0.15]{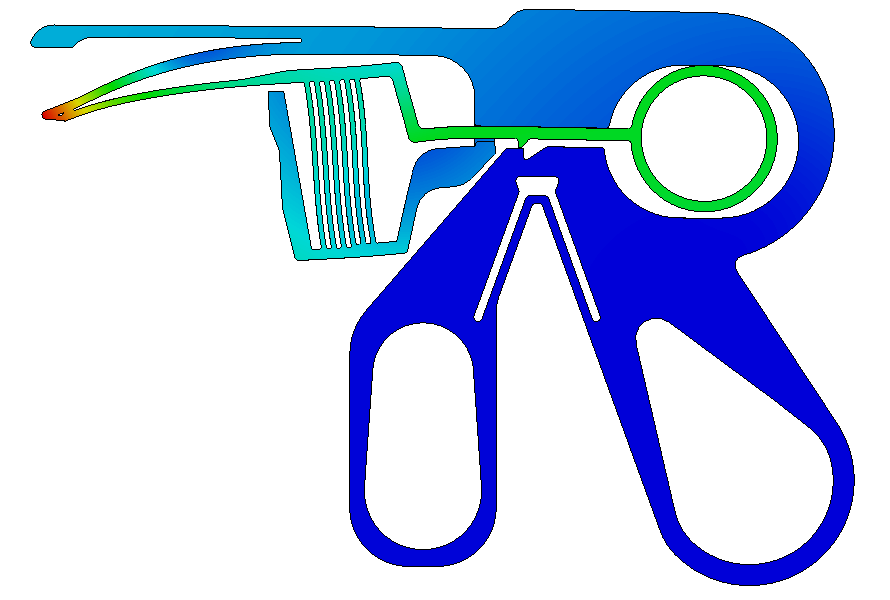}}             &     \includegraphics[angle=90, scale=0.023]{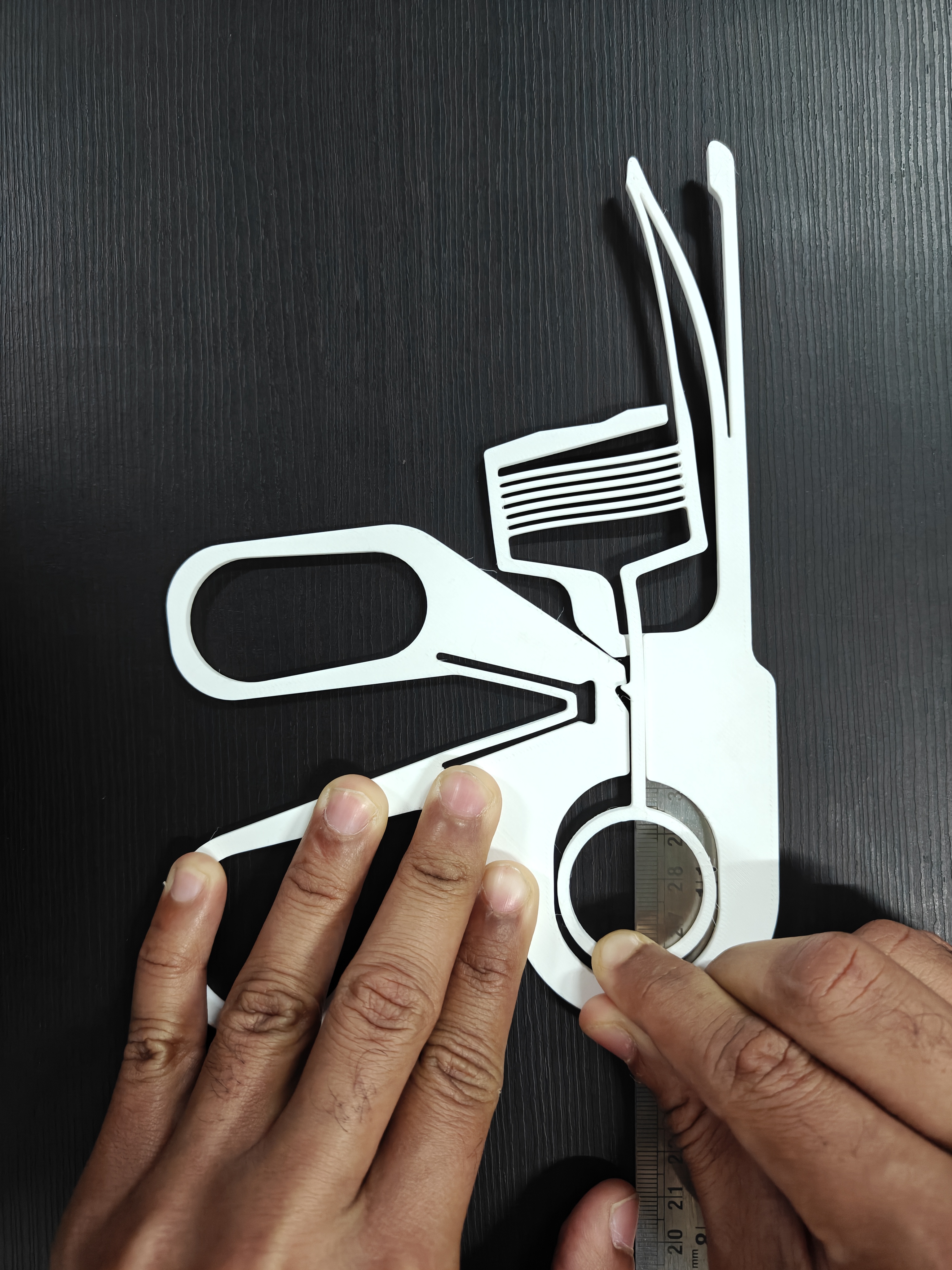}\\ \hline
 	7                                      & \multicolumn{1}{c|}{\includegraphics[scale=0.15]{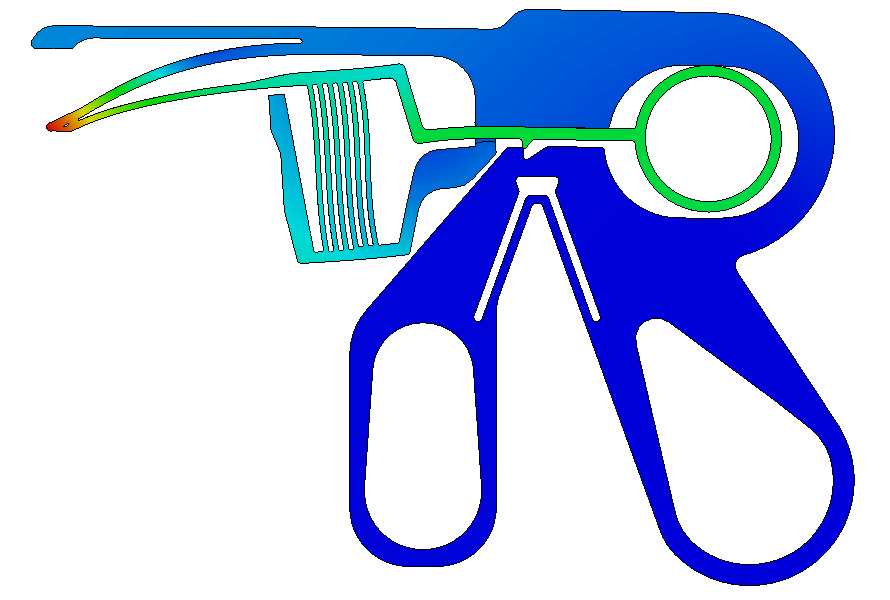}}             &     \includegraphics[angle=90, scale=0.023]{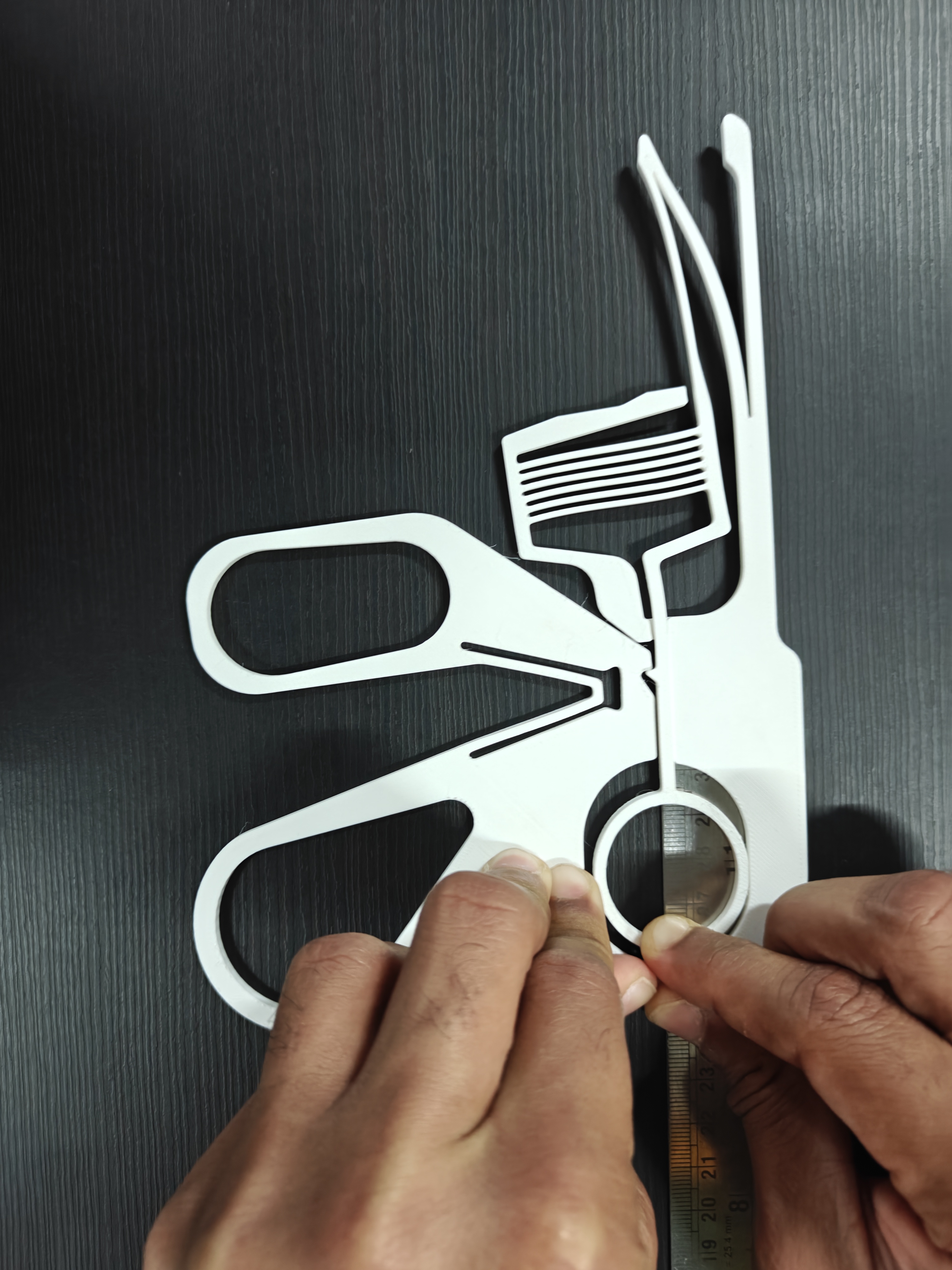}\\ \hline
 	8                                      & \multicolumn{1}{c|}{\includegraphics[scale=0.15]{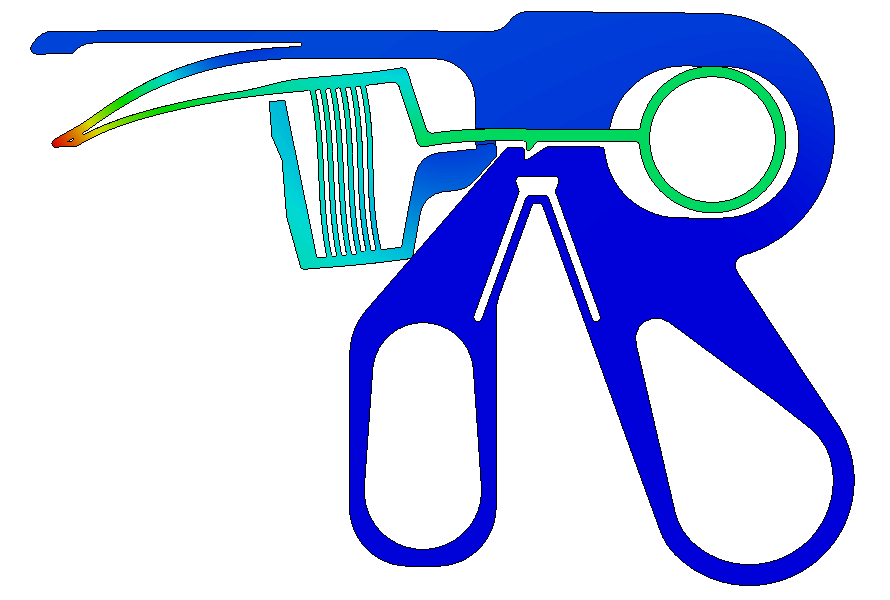}}             &     \includegraphics[angle=90, scale=0.023]{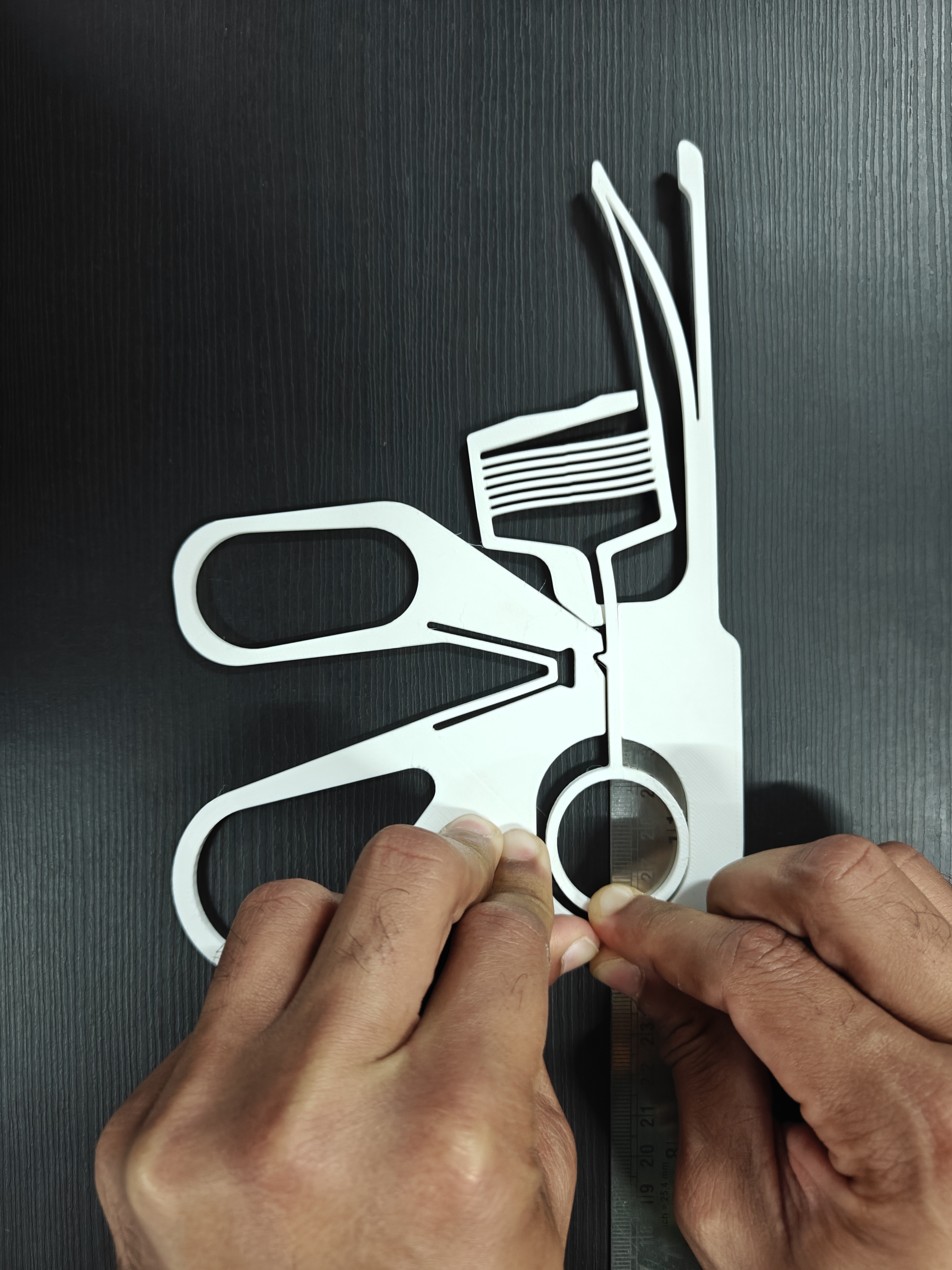}\\ \hline
 	\end{tabular}
 	\label{table:deformedprofiles}
 \end{table*}
 
 \begin{figure*}[h!]
 	\centering
 	\begin{subfigure}{0.19\textwidth}
 		\includegraphics[width=\linewidth]{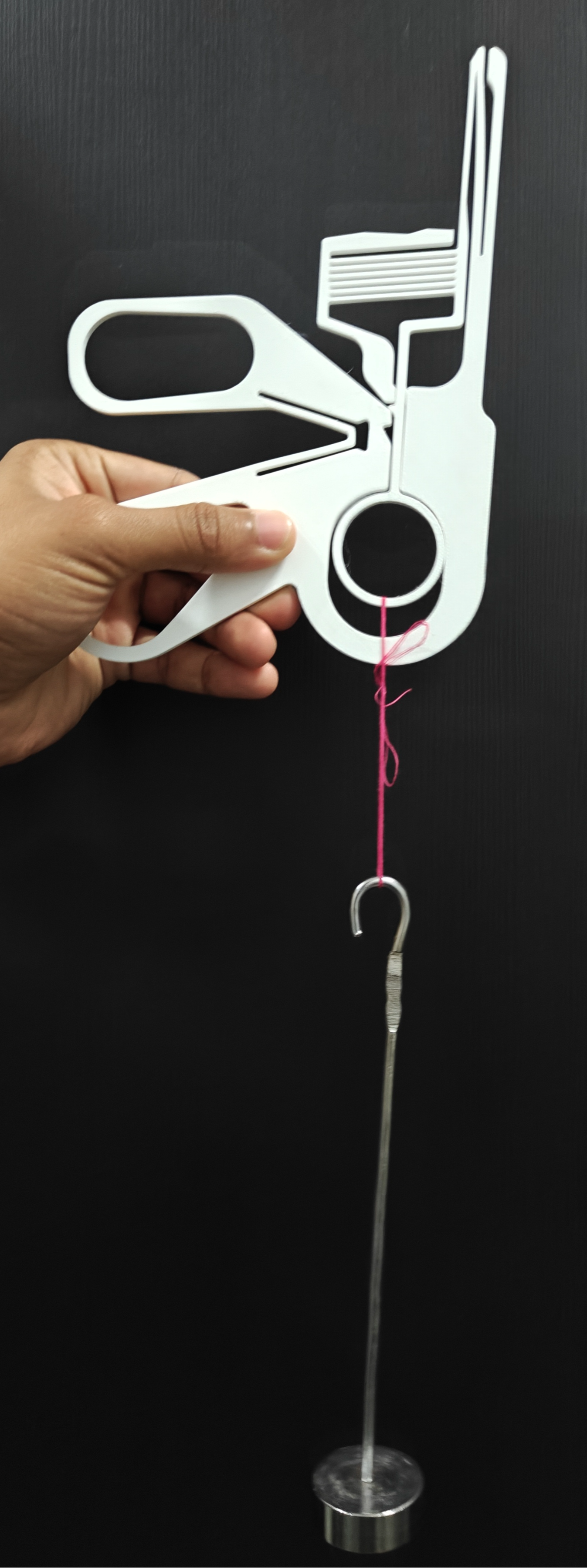}
 		\caption{}
 		\label{fig:exp1}
 	\end{subfigure}
 	\begin{subfigure}{0.19\textwidth}
 		\includegraphics[width=\linewidth]{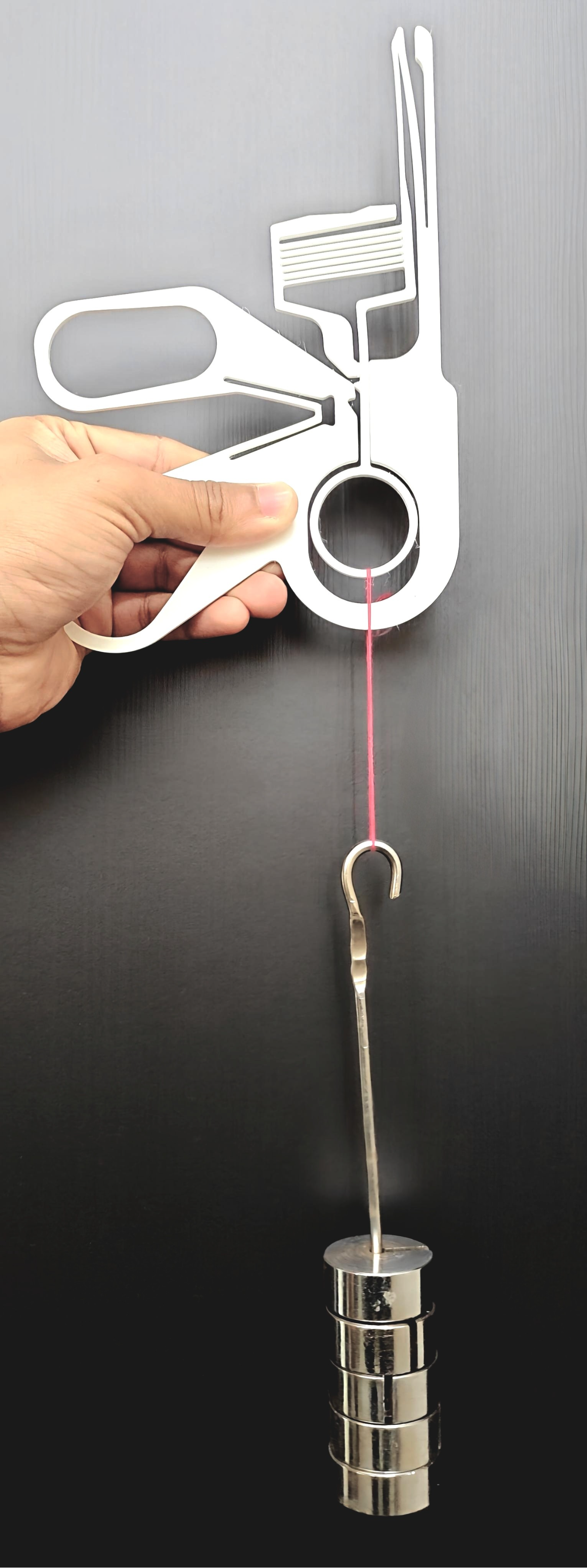}
 		\caption{}
 		\label{fig:exp2}
 	\end{subfigure}
 	\begin{subfigure}{0.19\textwidth}
 		\includegraphics[width=\linewidth]{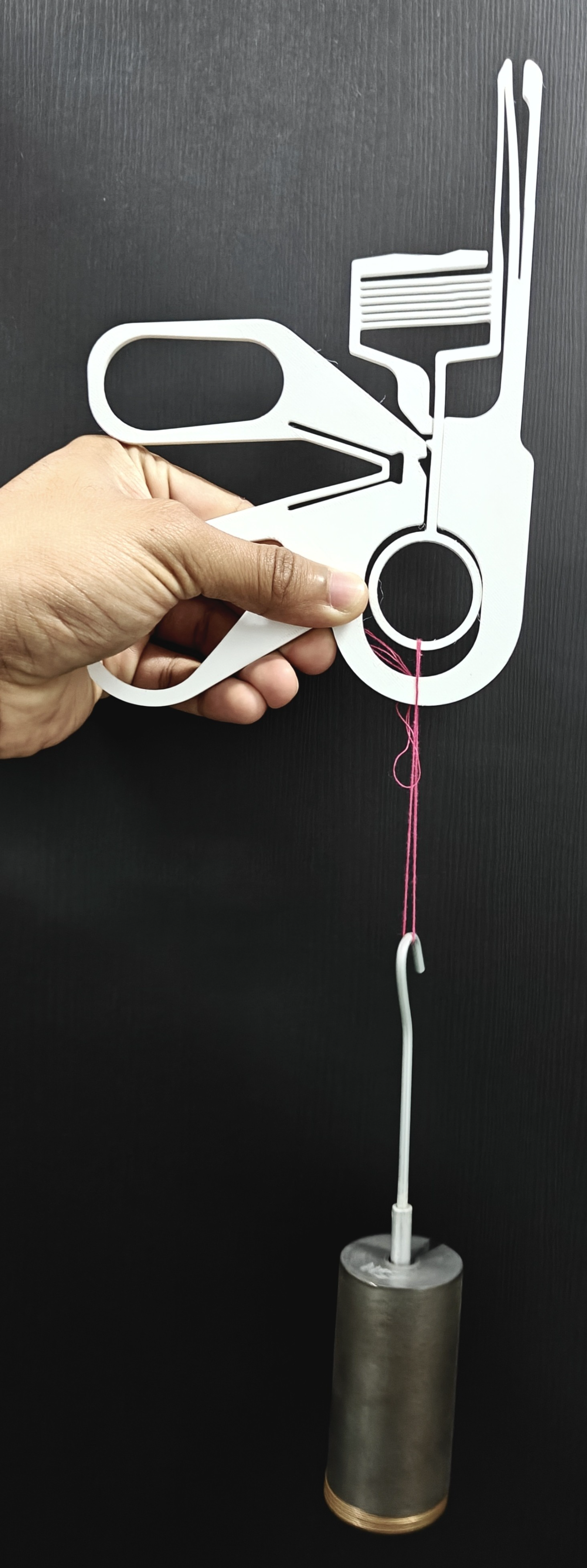}
 		\caption{}
 		\label{fig:exp3}
 	\end{subfigure}
 	\begin{subfigure}{0.19\textwidth}
 		\includegraphics[width=\linewidth]{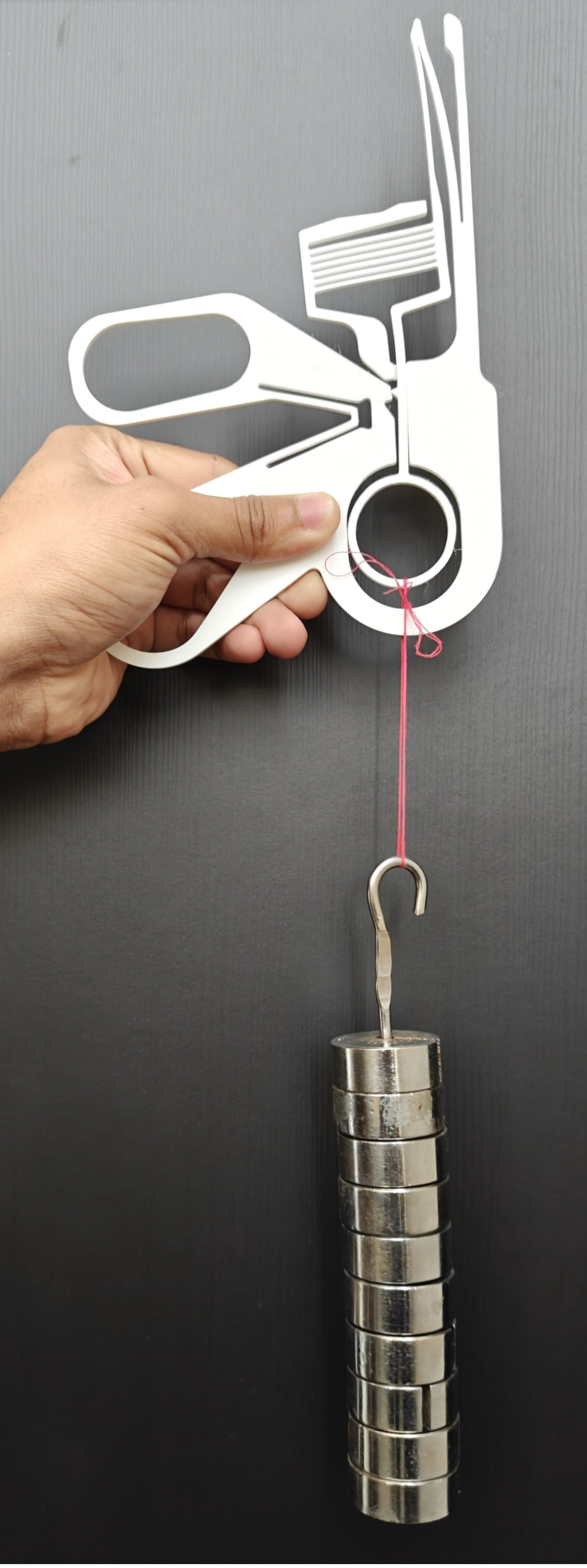}
 		\caption{}
 		\label{fig:exp4}
 	\end{subfigure}
 	\begin{subfigure}{0.19\textwidth}
 		\includegraphics[width=\linewidth]{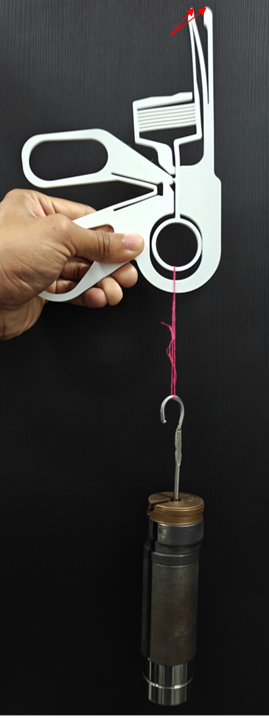}
 		\caption{}
 		\label{fig:exp5}
 	\end{subfigure}
 	\caption{The experimental setup used to measure the variation of jaw displacement versus force applied on the circular pull-back ring. The force is applied using masses of known weight, and the displacement is measured manually using an electronic vernier caliper. The marker in Fig.~\ref{fig:exp5} indicates the location of points where the vernier caliper measurements are taken to ensure correlation with the maximum displacement results obtained using FEM analysis.}
 	\label{fig:expsetup}
 \end{figure*}
 
 \begin{figure}[h!]
 	\centering\includegraphics[width=\linewidth]{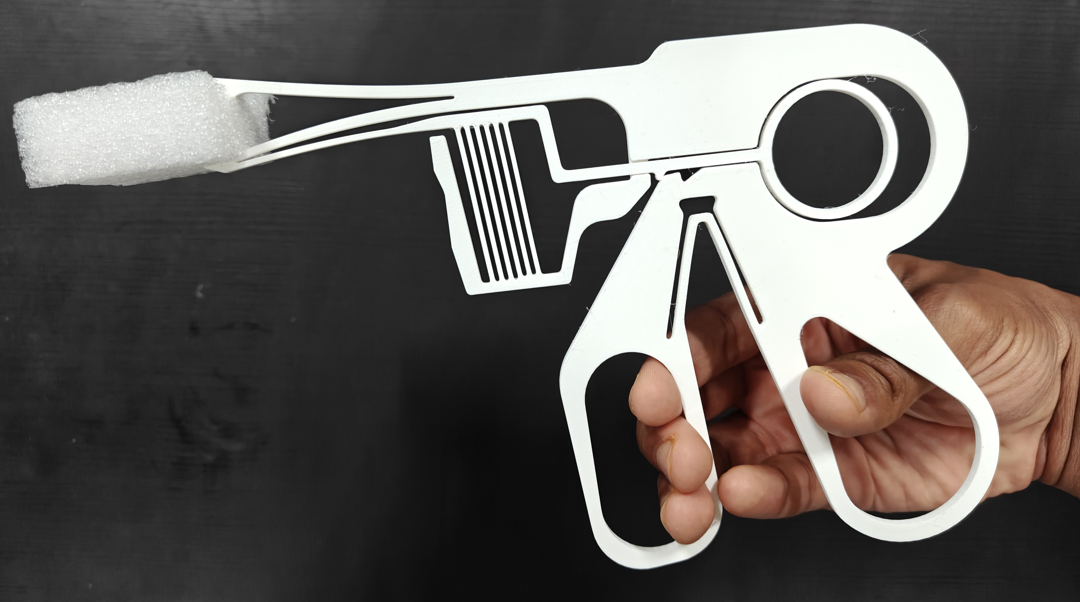}
 	\caption{The laparoscopic grasper in use, gripping a piece of packing foam.\label{fig:9-gripdemo}}
 \end{figure}
 \subsection{Compliant Gripper} \label{subsec:gripper}
 \hfill
 \par As mentioned in Sec.~\ref{sec:method}, a distinct compliant gripper is required for this particular actuation of the control push-rod; the gripper jaws must close when the control rod is pushed forward, and vice-versa. Further, because of the nature of the trigger-and-control push-rod mechanism, the gripper must be in its closed state when the mechanism is in its unstressed state. A similar compliant gripper mechanism is developed by Dearden et al.~\cite{Dearden}, which is based on an inverted L-arm gripper. This mechanism is not suitable for single-point actuation, as in this laparoscopic grasper. Guo and Tang~\cite{TOGripper} also developed a compliant single-point actuation gripper using their stiffness-oriented topology optimization method. The actuation forces of these grippers are too high, and their form factor does not match our application.
 
 \par Work in the area of single-point actuated compliant grippers has primarily been focused on symmetric grippers in which both jaws are actuated, either by a single, common actuator or multiple independent actuators~\cite{kumar2015topology,kumar2021topology,kumar2022topological, llanos2025systematic,KUMAR2026106531}. In this application of laparoscopic graspers, it is not necessary that both jaws move. Hence, we design our own tailored gripper intuitively, drawing inspiration from existing compliant grippers. 
 
 \par The gripper for our application is modeled as two separate jaws, where the upper jaw is fixed and connected to the main body of the laparoscopic grasper, and the lower jaw is connected to the control push-rod of the trigger mechanism (Fig.~\ref{fig:4}). The lower jaw is geometrically a cantilever beam at whose end a load is applied by this control push-rod, which is connected to the beam itself using a triangular extension. The 3D CAD model of the gripper end-effector is indicated in Fig.~\ref{fig:4}.
 
 \subsubsection{Finite Element Analysis \label{subsubsec:fem}} \hfill
 
 \par Nonlinear finite element analyses are performed to demonstrate the geometric properties of the proposed design in ABAQUS Standard using Static General steps. Meshing of the final design is performed using 15263, 10-node quadratic tetrahedral (C3D10) elements. Typically, such elements are compatible with the complex geometries used in the TiBCLaG and require no manual user intervention during meshing, unlike hexahedral elements, which may require user geometry modification. Second-order quadratic elements are preferred over linear elements because they can capture complex stress behavior, such as stress concentration and bending, more accurately. A displacement as a load is applied at the center of the circular pull-back ring to replicate the actual application. The load is applied in three steps: first, 3.2 mm; then, 6.4 mm; and finally, 8 mm. The final displacement of 8 mm is chosen so that the notch completely crosses its latch during the analysis. The intermediate values are set to 40\% and 80\% of the maximum displacement to capture the complex contact and snap-through behaviors. Multiple steps are used to carefully control the rate of load application and, more importantly, the stepping parameters for the complex contact interactions. The inside of the rear trigger handle is fixed using an \texttt{ENCASTRE} boundary condition, and the out-of-plane displacement and rotations are also arrested. Nonlinear geometry is turned \texttt{ON} to capture large deformations. The physical properties of the material are set as isotropic and elastic, with a Young's Modulus of 1800 MPa and a Poisson's Ratio of 0.3~\cite{PLA_Properties}. Although the fused filament fabrication process does not provide isotropic properties, it is not possible to precisely predict the final material properties in such processes due to the effects of fabrication errors, over- and under- extrusion, and infill percentages and patterns. Hence, a standard, simplified material description is used. Interactions are defined by hard contacts for normal behavior and frictionless contacts for tangential behavior to simplify the analysis. To validate the computational framework and ensure grid independence, a mesh convergence study is performed. Refining the mesh from 15263 elements to 38157 elements resulted in a maximum jaw displacement deviation of ~0.48\% and a peak von Mises stress deviation of about ~4.5\%. Since these deviations are comfortably within accepted computational tolerances, the base mesh of 15263 elements is utilized for all analyses to optimize computational efficiency.
 
 \par The analysis indicated that due to the large out-of-plane thickness compared to the in-plane thickness of the cantilever beam, large stresses are developed at its fixed end. This can be understood using a simple cantilever model where the maximum bending stress is inversely proportional to the area moment of inertia of the cross-section of the beam, and hence, its out-of-plane thickness:
 \begin{equation}
 	\sigma_{b, max} = \frac{M_b\ y_{max}}{I} = \frac{6 M_b }{b h^2}
 	\label{eqn7}
 \end{equation}
 where $\sigma_{b,max}$ is the maximum bending stress, $M_b$ is the bending moment, $I$ is the area moment of inertia of the cross-section, $y_{max}$ is the maximum transverse distance from the neutral axis of the beam, $b$ is the out-of-plane thickness of the beam, and $h$ is the in-plane thickness of the beam. Accordingly, the extruded thickness of the laparoscopic grasper is reduced to 3 mm. Moreover, to achieve the desired 20 mm opening of the jaws, it is observed that the displacement of the control push-rod at the gripper end remains insufficient. Hence, the number of beams is reduced iteratively to observe the effect on the displacement. Analysis indicated that six is the maximum number of beams that yield a 20 mm displacement of the jaw; any more would result in a decrease in displacement. It should also be noted that decreasing the number of beams beyond six would lead to a disproportionately large increase in stress. For example, a five-beam configuration led to an increase in stress by almost nine times (Fig.~\ref{fig:5v6BeamComparison}). Therefore, to find a proper trade-off between the number of beams, required deformation, and stress, we use six beams. Please note this is one of the ways to achieve the desired result.

 \par The highest stresses are observed at the out-of-plane hook between the support of the beams and the circular pull-back ring housing, i.e., the main body of the grasper behind the initial position of the beams. These large stresses are developed due to the arresting of out-of-plane deformation. Figure \ref{fig:midloads} indicates the geometry of the final laparoscopic grasper prototype during the retraction of the trigger at two different loads, and the values of stresses and displacements at these loads are summarized in Table~\ref{tab:2}.

 \section{Results and Discussion} \label{sec:results}
 This section provides fabrication specifications, experimental setup, results, discussion, and future work for the proposed TiBCLaG.
 
 \subsection{Fabrication of the Final Prototype} \label{subsec:finalfab} \hfill
 \par The final prototype of the laparoscopic grasper is fabricated in PLA (Fig.~\ref{fig:8-finalprototypes}) in full scale using the same printer settings as used previously. To ensure the prototype can be easily fabricated with a standard 3D printer, the grasper length is constrained to 200 mm. The total in-plane thickness of the end effector is also constrained to be compatible with a standard 10 mm trocar. It may be noted that industrial-grade laparoscopic graspers are manufactured in sizes of about 350 mm \cite{Dimensions-of-Laparoscopic-Graspers}. This prototype grasper can be extended to such lengths by simply increasing the length of the section between the gripper end-effector and the bending beams, i.e., the control push-rod and the fixed support of the cantilever beam. To maintain the displacement and stresses on the gripper during this modification, the geometry of the cantilever beam, including its length and thickness, must not be altered. However, preliminary analysis indicated that at such lengths, the support beams undergo a large deflection under actuation loads, which not only pose a medical hazard due to the risk of tissue entrapment, but also decrease the force transmission by absorbing the strain energy. Thus, the TiBCLaG is designed as a single-use mechanism that may be enclosed inside a reusable casing made of stainless steel or other suitable material. This casing will be dimensioned to operate within the limits of a 10~mm trocar and will provide the necessary stiffness to prevent unwanted bending of the support beams at these lengths.
 
 \par The adoption of such a reposable architecture also simplifies the question regarding fatigue life. The transition to a single-use mechanism ensures that the mechanical requirement is solely low-cycle fatigue, requiring reliable operation over typical procedure durations and actuation cycles. Moreover, sterilization requirements are simplified. While hospital-grade sterilization relies on autoclaving, this is incompatible with polymers like PLA. Hence, a single-use mechanism obviates the challenge of sterilization by ensuring that it is sterilized at manufacture using Ethylene Oxide (EtO) gas or gamma irradiation, which are suitable for medical grade PLA polymers \cite{perez2021sterilize}, or other similar, though slightly stiffer, materials like Polyoxymethylene or Polyether ether ketone. The reusable outer casing is sterilized using standard hospital-grade autoclaving.
 
 \subsection{Experimental Setups and Observations} \label{subsec:expsetup} \hfill
 \par A quantitative comparison between the kinematic performance of the grasper and the FEM result is indicated in Fig.~\ref{fig:comparison}. The experimental setup consisted of a ruler to measure the circular pull-back ring's displacement and a vernier caliper to measure the jaw's displacement. Table~\ref{table:deformedprofiles} presents the deformed profiles related to Fig.~\ref{fig:comparison}. The physical prototype shows very close agreement with the predicted displacements in the small-deflection regime (up to about 3.5 mm). For deflections greater than 4 mm, the fabricated prototype shows a clear divergence. At the maximum input pull-back ring displacement of 8 mm, the prototype gripper jaws displace by about 15 mm, compared to the 20.5 mm displacement expected from the analysis. One major reason could be out-of-plane deformation of the laparoscopic grasper, which was not accounted for in the finite element analysis using appropriate boundary conditions. To minimize this undesirable deflection, the stiffness in this direction must be increased by adding more material. Due to the nature of the fabrication process, however, the mechanism is designed to be as planar as possible. Utilizing other fabrication processes, such as stereolithography (SLA), injection molding or Laser Cutting, can improve out-of-plane stiffness without compromising manufacturability by seamlessly integrating complex out-of-plane geometries. Another important reason could be the description of the material's mechanical properties in ABAQUS. As mentioned above, fused filament fabrication does not provide isotropic properties; moreover, due to the effects of infill percentage, infill pattern, and fabrication errors, such as over- and under-extrusion, it is impossible to precisely predict the material's physical properties.
 
 \par The notch effectively restrains the motion of the control pushrod, maintaining the mechanism in its stressed bistable state. Pressing the trigger handle satisfactorily operates the trigger mechanism, pushing the control pushrod forward and closing the gripper end effector's jaws. 
 
 The experimental setup used to measure the force-displacement characteristics of the grasper is as follows. A non-elastic string is used to tie a loop around the circular pull-back ring. Varying forces are applied on the circular ring using masses of known weights placed on a cylindrical hook instrument of known weight. The displacement of the jaws is recorded using a vernier caliper at intervals of about 0.5 N until the mechanism is completely actuated, i.e., the latch settles into the provided notch cavity. Since gravity is being used to apply the load, the experiment is conducted with the grasper rotated clockwise by 90\textdegree~from its orientation during regular application. Figure \ref{fig:expsetup} presents a few loading cases. To ensure accurate correlation with the maximum displacement results obtained using FEM analysis, care is taken to ensure that the experimental displacement value recorded is the relative distance between the most distal points of the flat portion of the gripper jaws, without considering the added length due to the curved radius.
 
 
 \subsection{Challenges} \label{subsec:challenges} \hfill
 \par
\begin{figure}[h!]
 	\centering  
 	\begin{tikzpicture}[scale=1.5]
 		\begin{axis}[
 			xlabel={\large Force (N)},
 			ylabel={\large  Jaw displacement (mm)},
 			xmin=0,xmax=20,
 			grid=major,
 			legend style={at={(0.4,1)},anchor=north}
 			]
 			\addplot[smooth,{black}, line width=1.25pt, mark = star] 
 			coordinates {
 				(0.5,1.07)
 				(1,1.20)
 				(1.5,1.46)
 				(2,1.85)
 				(2.5,2.18)
 				(3,2.32)
 				(3.5,2.55)
 				(3.981,3.07)
 				(4.481,3.35)
 				(4.962,3.65)
 				(5.5,4.13)
 				(6,4.53)
 				(6.5,4.89)
 				(7,5.24)
 				(7.5,5.59)
 				(8,5.83)
 				(8.5,6.43)
 				(8.981,6.97)
 				(9.481,7.27)
 				(9.962,7.88)
 				(10.462,8.18)
 				(10.943,8.50)
 				(11.443,9.03)
 				(11.924,9.27)
 				(12.424,9.53)
 				(12.905,9.95)
 				(13.405,11.13)
 				(13.886,11.32)
 				(14.386,11.86)
 				(14.867,12.06)
 				(15.367,12.78)
 				(15.848,13.39)
 				(16.348,14.73)
 				(16.829,15.40)
 			}; \addlegendentry{\large Experimental displacement}
 		\end{axis}
 	\end{tikzpicture}
 	\caption{Variation of the gripper jaw displacement with the force applied to the circular pull-back ring of the laparoscopic grasper.}
 	\label{fig:forcevdispl}
 \end{figure}
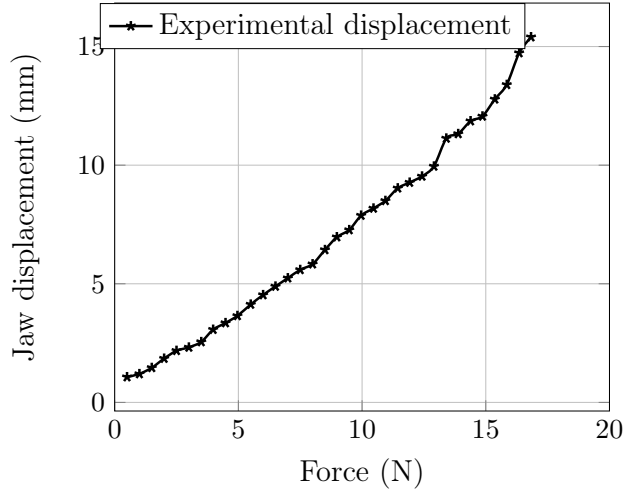

 The mechanism demonstrates reliable actuation and grasping (Fig.~\ref{fig:9-gripdemo}). However, since it is a compliant mechanism, the force applied on the circular pull-back ring would vary non-linearly with the displacement of the control pushrod, due to the storage of elastic strain energy in the mechanism. This is observed in Fig.~\ref{fig:forcevdispl}, where the slope varies according to the displacement of the gripper jaws, and hence, the displacement of the rings. The increase in slope at the end of the curve corresponds to the notch crossing. This non-linear variation can distort the force feedback to the operator, which may be one of the most significant drawbacks of applying compliant mechanisms to such applications. As a solution, future work will focus on developing mechanisms to ensure accurate force feedback transmission, which could include static balancing of the mechanisms. Initially, using the TEBC Model \cite{Liu2021ABC}, we design the grasper for a force magnitude of 20 N. It is observed that the force required to completely actuate the circular pull-back ring is 16.829 N, which is of the same order of magnitude. The variation in values may be attributed to the simplified assumptions of halving the forces for single I-beams and linearly adding the forces for multiple I-beams. 
 
 \par A quantitative comparison between the actuation forces and the jaw displacements of the TiBCLaG and conventional laparoscopic graspers reveals that the TiBCLaG requires higher actuation forces and lower jaw displacements than conventional laparoscopic graspers. While literature results provide inconsistent results regarding the actuation forces for conventional graspers, a typical range of 2 - 14~N is still slightly lower than the 16.829~N required to actuate the TiBCLaG \cite{olig2023output, van2012indirect, koc2016active, heijnsdijk2004influence}. The 15~mm jaw displacement obtained with the TiBCLaG may be suitable for grasping standard tissues, but it is still lower than the 22~mm reported for conventional laparoscopic graspers \cite{khan2024force}. However, previous literature on compliant laparoscopic grippers has demonstrated a jaw displacement of 10~mm \cite{Paper5}, which is significantly lower than that of the TiBCLaG, emphasizing the scope for improvement. The mechanical efficiency and force transmission ratio of the TiBCLaG are lower than those of conventional laparoscopic graspers due to the inherent nature of compliant mechanisms, which require increasing force as they are actuated. Improvement in such parameters may be pursued by using static balancing to cancel out the compliant mechanism's strain energy using a negative stiffness mechanism, thereby decreasing the actuation force required, and this is highlighted as a potential direction for future improvement.
 
 \par Another disadvantage is the jerky, impulsive force applied to the tissue when the surgeon actuates the trigger, which is caused by the large strain rate deformation of the flexible beams, releasing a tremendous amount of strain energy all at once. Further work must be done to alter the design of the mechanism to either absorb some part of this energy using some flexures or compliant springs or release the energy in increments. Adding multiple notches cannot only achieve incremental energy release but also provide the surgeon with greater control over tissue grasping. 
 
 \section{Conclusions}\label{sec:conclusion}
 This paper presents the design of a trigger-induced bistable compliant laparoscopic grasper (TiBCLaG) and its proof-of-concept validation. The presented hybrid design methodology, combining the TEBC model presented in previous literature for initial geometry estimation with FEM analyses and iterative physical prototyping to refine the geometry and verify the performance, offers a framework for monolithic CM laparoscopic grasper design. Developing a monolithic, fully compliant laparoscopic grasper reduces the part count, significantly decreasing manufacturing and assembly costs. The TEBC Model is used as an initial design framework for the grasper's trigger mechanism. This initial design is modified through multiple prototyping iterations to yield the final trigger mechanism. The compliant gripper mechanism is then intuitively designed, drawing inspiration from existing compliant grippers, and integrated into the trigger mechanism with slight modifications. The design is initially validated through finite element analysis and then fabricated using fused deposition modeling. 
 
 Future work will focus on the kinetic validation of the TiBCLaG and on a thorough comparison with conventional laparoscopic graspers. Iterative geometry tuning may be required to achieve the required grasping forces. The development of a reusable casing, along with the investigation of a suitable material, should also be considered. Further development of the design can be pursued in multiple ways. Integrating multiple notches into the design, rather than a single notch, to restrict strain energy release can provide the surgeon with greater control when grasping tissues. Static balancing of both the trigger and the gripper mechanisms can improve the force perception by providing accurate haptic feedback.

 Since fused deposition modeling presents inherent print-to-print variability that limits rigorous physical testing of the prototype, using alternative fabrication methods like plastic injection molding is key in future iterations, as it may not only improve correlation between analysis and testing but also allow for geometric modifications that enhance the mechanism's out-of-plane stiffness, creating a more efficient energy transfer by preventing losses in undesirable directions. Employing such fabrication methods to manufacture the single-use plastic mechanism for eventual medical deployment significantly reduces manufacturing costs, and the monolithic nature of the mechanism drastically reduces assembly costs, a major advantage of the TiBCLaG. While preliminary testing is performed on foam, subsequent validation must transition to ex vivo biological tissue to accurately assess grip security and tissue-tool interactions. This biological testing may prove particularly critical in evaluating the efficacy of the TiBCLaG in delivering a smooth strain-energy release, possibly utilizing the integrated multiple notches as mentioned above. The success of this investigation reveals the potential to develop such advantageous mechanisms to replace industrial tools, while the highlighted refinements can enhance operator comfort. 
 
 \section*{Acknowledgment}
 The authors thank A. Padmaprabhan from the Department of Mechanical and Aerospace Engineering at IIT Hyderabad for his valuable contributions during the development phase of the mechanism. They thank Santosh Bharghav for his insightful suggestions at the Students Mechanism Design Contest, 2025. They are also grateful to Just L. Herder, TU Delft, the Netherlands, for his stimulating conversation at the 7th International and 22nd National Conference on Machines and Mechanisms (iNaCoMM) 2025. PK thanks Anusandhan National Research Foundation, India, for the support under the MATRICS project file number MTR/2023/000524. 
 
 \section*{Conflict of Interest}
 The authors declare no conflicts of interest or competing financial interests.

  \bibliographystyle{ieeetr}
  
  \bibliography{bibliography}
  
\end{document}